# Metaverse for Healthcare: A Survey on Potential Applications, Challenges and Future Directions

RAJESWARI CHENGODEN[1], NANCY VICTOR[1], THIEN HUYNH-THE [2], (Member, IEEE), GOKUL YENDURI[1], RUTVIJ H.JHAVERI[3], (Senior Member, IEEE), MAMOUN ALAZAB[4], (Senior Member, IEEE), SWETA BHATTACHARYA[1], PAWAN HEGDE[1], PRAVEEN KUMAR REDDY MADDIKUNTA[1], and THIPPA REDDY GADEKALLU[1], (Senior Member, IEEE)

[1]School of Information Technology and Engineering, Vellore Institute of Technology, Tamil Nadu, India (e-mail: rajeswari.c, nancyvictor, gokul.yenduri, sweta.b,pawan.hegde, praveenkumarreddy, thippareddy.g @vit.ac.in)

[2]Department of Computer and Communication Engineering, Ho Chi Minh City University of Technology and Education, Ho Chi Minh, 71307, Vietnam (e-mail: thien.huynh-the.2016@ieee.org)

[3]Department of Computer Science and Engineering, School of Technology, Pandit Deendayal Energy University, Gujarat, India (e-mail: rutvij.jhaveri@sot.pdpu.ac.in)

[4]College of Engineering, IT and Environment, Charles Darwin University, Casuarina, NT 0909, Australia (e-mail: alazab.m@ieee.org)

Corresponding author: Mamon Alazab (e-mail: alazab.m@ieee.org), Thippa Reddy Gadekallu (e-mail:thippareddy.g@vit.ac.in).

**ABSTRACT** The rapid progress in digitalization and automation have led to an accelerated growth in healthcare, generating novel models that are creating new channels for rendering treatment with reduced cost. The Metaverse is an emerging technology in the digital space which has huge potential in healthcare, enabling realistic experiences to the patients as well as the medical practitioners. The Metaverse is a confluence of multiple enabling technologies such as artificial intelligence, virtual reality, augmented reality, internet of medical devices, robotics, quantum computing, etc. through which new directions for providing quality healthcare treatment and services can be explored. The amalgamation of these technologies ensures immersive, intimate and personalized patient care. It also provides adaptive intelligent solutions that eliminates the barriers between healthcare providers and receivers. This article provides a comprehensive review of the Metaverse for healthcare, emphasizing on the state of the art, the enabling technologies for adopting the Metaverse for healthcare, the potential applications and the related projects. The issues in the adaptation of the Metaverse for healthcare applications are also identified and the plausible solutions are highlighted as part of future research directions.

## I. INTRODUCTION

Healthcare is one of the most significant determinants of ensuring general, physical, social and mental well-being of the entire human population in the world. The primary objective of any healthcare system is to channelize its efforts towards activities that promotes, restores, maintains and improves healthcare services. It also contributes immensely towards efficient development of a country's economy and industrialization. This sector has thus observed rapid growth and revolution being highly exposed to technological evolution in order to enhance the experience of interaction caregivers, patients and related stake holders. The revolutionizing of digital healthcare has acted as a key enabler of change in the healthcare industry [1]. The initiation of digital health services using digital and internet tools has impacted the interaction between patient-physician at a very large scale wherein changes were visualized due to technologies such as blockchain, augmented reality, virtual reality [2]. Despite the rapid progress in the healthcare sector, certain issues in this sector remains to be prominent such as insurmountable pressure of long-term chronic ailments, accelerating costs, aging population, insufficient healthcare workforce and availability of limited resources. These predominant issues have instigated the need to fetch healthcare services to the living room of individuals [3], [4]. The recent pandemic of COVID-19 have added enormous pressure to the global healthcare sector and related workforce, infrastructure and supply chain management. The COVID-19 has been the primary reason for accelerating rapid change across the healthcare ecosystem and have compelled the stake holders to pursue adaptation and innovation of all the technologies used in this sector [5], [6].

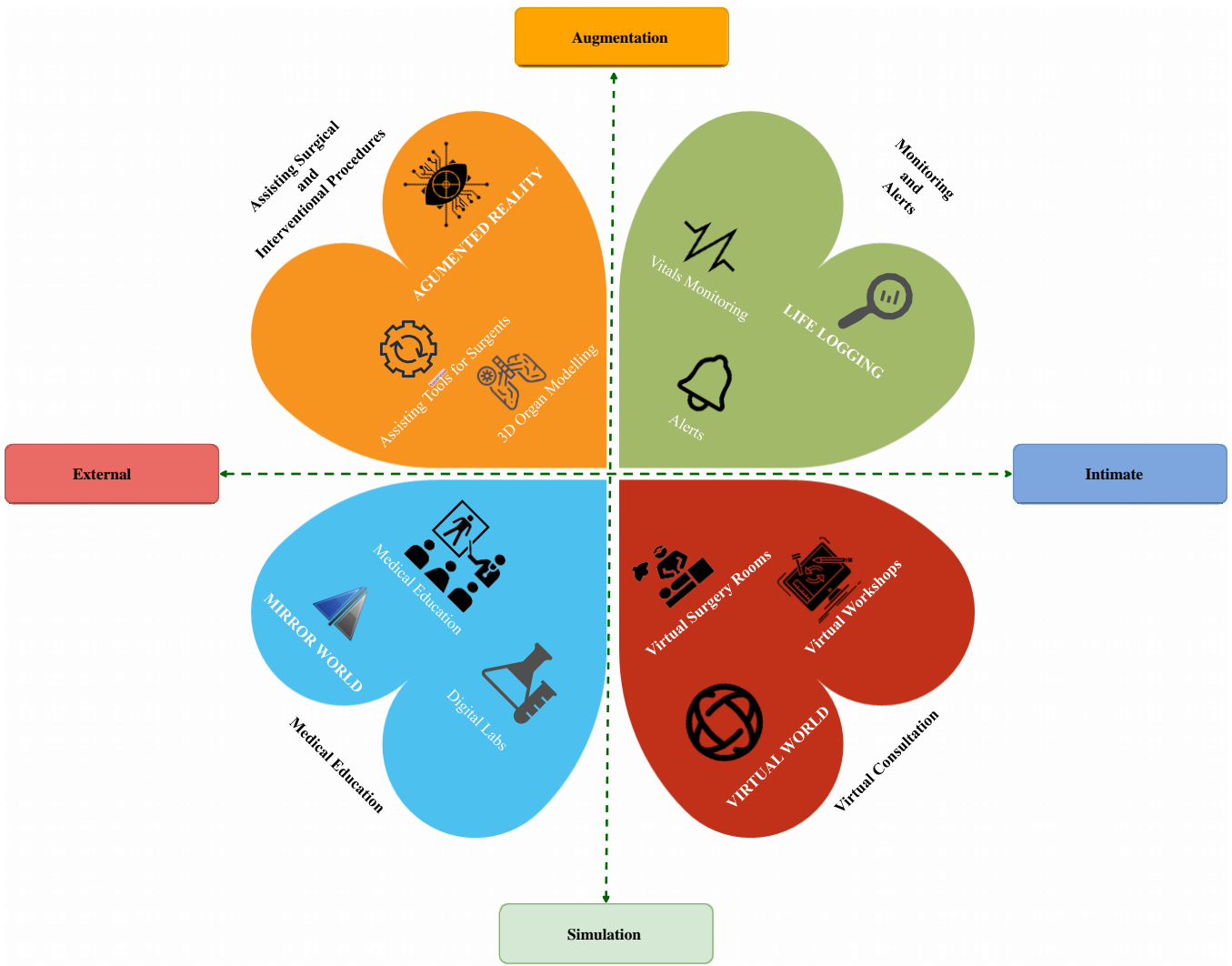

**FIGURE 1.** The Four Axis of the Metaverse Implementation in Healthcare

The post pandemic era has resulted in major foundational shifts in the healthcare domain. As an example, the present generation consumers have started taking active participation in healthcare related decision making followed by enthusiastic adoption of virtual healthcare systems and related digital innovations. There has also been active impetus for the use of inter-operable data and data analytics; unprecedented collaborations in therapeutics development which has compelled governments, healthcare providers and other stake holders to adapt and innovate [7]. But there still exists prominent challenges and the response to these challenges would pave the path for the journey to the future of healthcare sector. Consumers (patients) and their experiences are the accelerating force as their ever evolving requirements and goals drive innovations in this sector. Their primary preferences include development of digitally enabled, on-demand and seamless patient-clinician interaction ensuring delivery of patient centric services across geographical boundaries and socio-economic clusters [8]. Each individual patients' health journey stands unique and it is imperative to acknowledge this fact, thereby customize the specific services and elevate each interaction to the level of personalized healthcare experience [9]. The need to deploy advanced digital tools and services has become a necessity in order to provide optimized consumer satisfaction, enabling tracking and monitoring of health status and improved adherence to medications. The healthcare consumers are progressively keen on sharing their confidential data and hence the need for organizations has evolved in order to provide interoperability between organizations and retain consumers trust demonstrating reliability, transparency and empathy in their operation [10]. The present expectation is to shift the focus from healthcare to health and well being motivating changes in design of service offerings and delivery channels. The organizations are thus encour-aging implementation of virtual-care, remote monitoring, digital diagnostics and decision support systems, at-home prescription delivery systems and self-service applications for education and social support. This digital transformation has significantly impacted the healthcare ecosystem by improving their working capability, access to services, patient-clinician experience by using artificial intelligence, cloud computing, Augmented Reality (AR), Virtual Reality (VR) technologies [7].

The healthcare system in the Metaverse provides healthcare service experience that is interactive, immersive and recreational customized to meet individual patients needs. The Metaverse comprises of advanced technology revolutions namely artificial intelligence, augmented reality (AR), virtual reality (VR), telepresence, digital twinning and blockchain which has huge impact on healthcare [11]. The use of these technologies provides exposure towards new ways of delivering treatment in significantly lower cost thereby enhancing patient outcomes. The Metaverse creates the experience of virtual world using internet wherein the human gestures and emotions get simulated. It involves the complete social and economic structure of both real and virtual world [12]. The four axis of the Metaverse implementations in healthcare is presented in Fig. 1

The Metaverse technologies can help healthcare professionals in effective planning and diagnosis of diseases [13]. In 2020, the neurosurgeons in the Johns Hopkins Hopsital performed a surgery using augmented reality headset developed by Augmedics. The procedure involved use of a see-through eye display that projected patients' anatomy images similar to an X-ray vision for fusing six vertebrae in the patients' spine to relieve constant back pain [14], [15]. The Metaverse environment enables enhanced surgical pre-operative planning by transforming CT scans into 3D reconstructions using headsets. This also helps the surgeons to specifically view, isolate and manipulate anatomical regions to perform critical surgeries [16]. The Metaverse tools also provide amplified prescription treatments. As an example, EaseVR is a prescription-based solution that uses cognitive behavioral therapies to treat patients suffering from backpain using VR headset and controllers. These tools help in providing deep relaxation, attention-shifting and interoceptive awareness that cater to the physiological aspects of pain. Plastic surgery is an extremely complex procedure which requires reconstruction of human body parts [17], [18]. In case of plastic surgeries the use of VR in the Metaverse could play an important role wherein the virtual avatar could accurately predict the outcome of a plausible plastic surgery. The Metaverse surgeries require detailed understanding of the human anatomy and ability to use instruments with higher dexterous grasping capability which are flexible and adjustable on an individual basis. The potential of its huge ranges from simple procedures to complex surgeries including removal of tumors and performing of intricate and complicated spine surgeries [19]. The Metaverse in the radiology domain has the potential to unleash advanced capabilities in image visualization enabling radiologists to view dynamic images in more details resulting in enhanced diagnosis and accurate decision making. Also, it would provide opportunity for better training in radiology and ability to collaboratively work on 3D medical images while being located at different geographic locations [20]. The healthcare metaverse could improve patients engagement with the help of high-quality immersive content and features of gamification aiding clinicians to explain complex concepts to patients, provide walk-throughs on procedures they would undergo and ensure patients take accurate prescribed medications.The use of digital twins solutions in The Metaverse will keep the consumers well informed and engaged on their treatment wherein the patient's vitals, CT scans, health records and genetic test results are integrated to develop a digital simulation of the patients anatomy and physiology to monitor and generate insights on their health condition [21]. The health data can be visualized by the patients on the virtual dashboard helping them to communicate with clinicians, researchers, nutritionists and other stake holders for achieve individual care and treatment [20].

The recent pandemic has generated the need for remote healthcare services and the Metaverse has the potential to provide greater experience than the classical videoconferencing based telemedicine systems [20]. The Metaverse systems could use AR glasses to access live videos and audio communications for communicating with physicians in real-time. The AR solutions would enable the respondents to interface directly and also provide live streaming of emergency situations to off site clinicians to provide accelerated and on-time, on-spot treatment [22].

The Metaverse has the potential to revolutionize medical education and training. The use of AR provides the conducive environment to explain practical procedures rather than disseminating theoretical knowledge. The reputed institutions are slowly moving towards the use of VR based technology, AR, mixed reality (MR) and AI to educate doctors by simulating complex real-time procedures and providing information on the cellular level details of the human anatomy [23]. The potential applications of the Metaverse in healthcare could be categorized into 4 categories depending the on the enabling technologies being used namely AR, life logging, VR and mirror world. As an example, an AR T-shirt would allow students to get clear visualization of the human body in an anatomy lab. In case of mirror world implementation, it would reflect the real world intact in association to the integration of data pertinent to environmental information using virtual maps and modeling techniques. The use of the Metaverse was further enhanced due to the recent pandemic situation. In a classical mirror world platform, individuals located at distant location meet and play games in the mirror world. Similar idea in the healthcare sector allows participants in the scientific community to contribute in scientific research through games. As an example, the team of David Baker at the University of Washington study protein structures and use digital labs for enabling researchers fold protein amino acid chains. As part of the game, when the protrusion structure matches, the player is awarded points and progresses to the next level. thus, these advanced technologies provide more refined visualizations, understanding and practise of new techniques. The Metaverse has the ability to provide 360 degrees visualization of the body ailments and can act as the most helpful surgical training tool fostering optimum level of cooperation and highest degree of immersion. The technology although is at the stage of experimentation but when implemented it could deliver great outcomes in the field of healthcare education and training [24].

The global statistics of the Metaverse reveal that the global healthcare market in the Metaverse holds a value of 5.06 Billion Dollars in 2021 and is expected to reach 71.97 Billion Dollars by 2030 with a compound annual growth rate (CAGR) of 34.8 percent during considered period of forecast (2022-2030) [25]. North America is expected to dominate the other countries during the said period due to the strong presence of the Metaverse-focused companies in this region. There is also existence of strong infrastructure, integration of AR-VR devices and platforms in their healthcare sector which have attracted more investments in AR based products and applications and initiated similar updates in their software - hardware infrastructure [25].

The potential of the Metaverse has been explored by various researchers. The study by [26] discussed the potential contribution of the Metaverse in transforming and enhancing healthcare in terms of collaborative working, education, clinical care, monetization and wellness. The study by [27] discussed the role of smart health intelligent healthcare system in the Metaverse wherein various case studies of AI and data science applications in healthcare are explored in case of hospital administration. The review of the case studies unravel various challenges in the application of the Metaverse and related applications in healthcare. In [28], the potential of smart healthcare 5.0 is explored which contributes in ambient tracking of patients, provides emotive telemedicine, performs telesurgery, wellness monitoring, conducts virtual clinics and renders personalized care. The article discusses the plausible role of the Metaverse in leveraging digital connectivity through enhanced healthcare experience in virtual environments. Although the components of the Metaverse are decentralized, the use of blockchain induces transparency and immutability in the stored transaction of the Metaverse applications. The study investigates the possibility of using telesurgical scheme involving patients, doctors in a virtual hospital setup using the Metaverse technology. The study in [29] explored the scope of the Metaverse applications integrated with IoT, blockchain, AI and other relevant technologies. The article reviewed the various enabling technologies metaverse uses to unveil its full potential applications in the healthcare sector. The study in [30] explored the potential of the Metaverse applications as part of Medical Internet of Things (MIoT) using AR and VR glasses. The study involved participation of a panel of doctors who suggested the feasibility of the Metaverse implementations based on three basic MIoT functionalities namely comprehensive perception, reliable transmission and intelligent processing using the Metaverse platform. The frameworks constituted of AR/VR glasses integrated with holographic construction, emulation, VR integration and interconnection. The study in [31] presented a brief overview on the Metaverse and its potential to revolutionize healthcare in association to its implementation challenges of high cost, loss of privacy, ethical concerns and agreement issues from healthcare administrations and organizations. The contribution of recent studies in the Metaverse and the unique contribution of this review paper is presented in Table 1.

## II. STATE OF THE ART: EXISTING DIGITAL AND SMART HEALTHCARE ENABLING TECHNOLOGIES

This section conveys the state-of-the-art technologies enabling the digital and smart healthcare. They includes sensor, big data, artificial intelligence (AI), wireless communication networks, Internet of Things (IoT), edge and cloud computing, and immersive technology.

### A. SENSORS

The precision of personal healthcare analysis partly depends on the measurement of different physiological parameters using sensors. Despite being readily available, medical equipment is costly and consumes a lot of power. To this end, sensors are purposefully integrated in IoT-based medical systems, in which they can measure different biological parameters and monitor real-time patient healthcare-related data, especially when being uplifted to the remote sensing technology. With biomedical sensors, medical systems are flexible with machine-to-machine interactions, saving time of both patients and medical institutions, and offering treatment plans over telediagnosis. Among various sensors, temperature, electrocardiogram (ECG), and pulse are the most important ones for health status evaluation besides blood pressure, accelerometers, and imaging sensors. The following part will discuss some relevant sensors commonly used in healthcare and medical systems.

Compared with the digital temperature sensors, the analogues ones are not frequently used in recent healthcare devices and systems because of its requirement of external or internal ADC (analog-to-digital converter) component. Some digital temperature sensors are usually used as built-in component in wearable patient monitoring systems to measure and monitor patient's body temperature [32]. Recently, an innovative wearable temperature sensor technology which utilizes freestanding single reduction graphene oxide fiber was introduced and developed with many benefits, such as fast response, high stability and repeatability under mechanical deformation, and wearable comfort [33]. This wearable, freestanding, and fiber-based temperature sensors is promising for healthcare measurement and monitoring applications. ECG, a sensor to measure heartbeat and strength, is widely used for many heart disease diagnosis applications and services [34], which can be found in connected wear-able devices, such as smartwatch and specialized shimmer ECG units. For example, in [35], the ECG data, collected by different portable devices and from different healthcare institutions, was analyzed to accurately detect heart abnormalities in the real-time conditions using advanced machine learning (ML) algorithms. Beside heart state monitoring, ECG is also applied for some sophisticated cardiovascular diagnoses, such as atrial fibrillation, myocardial ischemia, and myocardial infarction. Different with ECG, pulse rate sensor is used to verify how well the heart is working, especially in emergency scenarios of cause determination. In addition, some other sensor types have been exploited in the healthcare domain: pulse oximeter is to measure human blood oxygen saturation and detect arrhythmia, blood pressure sensor as a non-invasive sensor is to measure human blood pressure (can be extended for systolic, diastolic and mean arterial pressure measurement using the oscillometric method), and blood glucose monitors and meters to measure glucose levels of diabetic patients [36]. Nowadays medical imaging plays a considerable role in the healthcare and medical domains with different techniques, such as X-ray computerized tomography, magnetic resonance imaging, ophthalmology, optical coherence tomography, and positron emission tomography [37], already commercialized in medical scanner systems.

TABLE 1. Comparison and Contributions

| Ref. | Application of the Metaverse in Healthcare | Discussion on Key Enabling Technologies | Projects and Real-time case studies in Healthcare |
|------|---|---|---|
| [26] | ✓ | X | X |
| [27] | ✓ | X | X |
| [28] | ✓ | X | X |
| [29] | ✓ | X | X |
| [30] | ✓ | X | X |
| [31] | ✓ | X | X |

### B. BIG DATA

In general, big data is currently characterized by seven Vs as follows: volume (a large amount of data), variety (including structured, semi-structured, and unstructured data with different formats), velocity (high rates of data inflow and real-time processing), veracity (detailed data accumulation), value (in-depth and meaningful information), variability (offering extensionality and scalability), and valence (data interconnection). About the big data workflow, a four-step process should be followed with big data generation (generated from different sources: internal data from hospitals and medical centers, IoT data, data from Internet, and biomedical data), big data acquisition (including data collection, transmission, and pre-processing), big data storage, and big data analysis.

In the last decade, several modern technologies and tools based on distributed architectures, along with large memory and powerful computing units, have been introduced for big data processing in healthcare and medical domains:

- Big data collection: Sqoop is an open-source framework for data import and export between Hadoop Distributed File System (HDFS). Flume is a reliable and distributed open-source service for assembling, aggregating, and transferring batch files, log files, and streaming data from healthcare systems to HDFS for storage.
- Big data processing: MapReduce is a programmable model that simplifies and accelerates a massive amount of data, featured by three main functions (e.g., mapping, sorting, and reducing) and a master-slave architecture. YARN (Yet Another Resource Negotiator) is more superior than HDFS with dynamic allocation of system resources to optimize the resource utilization efficiency. Some other open-source big data processing frameworks include Storm, Flink, Spark, and Zookeeper.
- Big data storage: Hadoop performs the distributed system HDFS and the non-relational databased HBase for big data storage. HDFS, a primary component of a Hadoop cluster, provides high throughput access to application data and is suitable for high-volume data applications. As the advantages, HDFS are highly fault-tolerant and assists low-cost hardware. HBase, which is an open-source platform built on top of HDFS for low-latency operation, is flexible, distributed, and scalable with real-time querying and configurable data partitioning.
- Big data analysis: There are some open-source platforms for big data analysis, such as Pig, Hive, and Mahout. Pig, introduced by Yahoo to analyze big data flows, is an open-source big data analysis platform with several strong points, such as easy programming, semi-structured and unstructured data supporting, massive amounts of data processing. Hive, created by Facebook, is a data warehouse system to facilitate Hadoop implementation. Although featured by strong interactive interface with numerous built-in functions for data analysis, Hive mostly makes easier with structured data instead of others, thus being unfriendly to healthcare and medical data. Mahout is an open-source library with built-in functions for data mining and machine learning.

### C. ARTIFICIAL INTELLIGENCE

AI in healthcare is an all-encompassing term used to describe the use of ML algorithms to mimic human cognition in presenting, analyzing, understanding, and learning complex healthcare and medical data. ML, a subfield of AI, is a set of statistical algorithms that can learn from data via a training stage and make prediction via an inference stage. By giving the computer the ability of learning pattern from data automatically without explicit and manual programming, ML is therefore used in a wide range of scientific domains, including the healthcare and medical with precision medicine (i.e., predicting and recommending an optimal treatment strategy to succeed on a patient based the analysis and diagnosis of multiple patient attributes).

In fundamental, most of the existing AI/ML algorithms can be grouped into two categories: traditional techniques and advanced techniques, which can solve three principal problems: clustering, classification, and regression. Conventional AI/ML algorithms usually performs four data-based learning types [38], [39]: supervised learning (i.e., learn the relation between input and output via a mapping function using labeled data and classify/predict the outcome for an unforeseen input sample using the trained model), unsupervised learning (i.e, involve the utilization of ML algorithms for unlabeled data analysis and clustering, and can find out data groups without the need for human intervention), semi-supervised learning (i.e., trained upon the combination of clustering similar data using an unsupervised learning algorithm and using the existing labeled data to label the remaining unlabeled data), and reinforcement learning (i.e., make a sequence of decisions, in which an agent learns to attain a goal in an uncertain and complex environment). Recently, deep learning (DL) is a subset of ML with advanced architectures, e.g., recurrent neural network, long short-term memory network, and convolutional neural network, relying on multi-layered artificial neural networks, to attain groundbreaking performance in many classification and regression tasks of healthcare and medical domains [40]. Unlike traditional ML techniques, DL can automatically learn underlying features of unstructured data without human intervention or human domain knowledge [41]–[44]. The highly flexible architectures of DL allow learning systems to process raw data directly and improve learning performance when the data is provided enough [45]. Some well-known deep architectures include: recurrent neural network – RNN (this architecture has some feedback connections associated with the preceding layers to maintain memory of past inputs and process models in time), convolutional neural network – CNN (as one of the most successful deep network architectures, this architecture leverages principles from linear algebra to identify complex patterns from high-dimensional unstructured data), self-organizing map (this unsupervised neural network is to find clusters of the input data points by reducing its dimensionality), and autoencoders (this special type of neural network is trained with the compression and decompression functions to map its input to output).

Natural language processing (NLP), rule-based expert system, and physical robot have played some key roles in the modern healthcare and medical domains, especially when being strengthened by ML algorithms and DL architectures [46]. NLP with statistic and semantic approaches can analyze unstructured clinical notes, prepare explanation reports (such as radiology and physical examinations), transcribe patient interactions, and guide conversational AI. Rule-based expert systems based on a collection of "if-then" rules manually built for "clinical decision support", which require human experts and knowledge engineers for rule construction in a specific knowledge domain. Despite being more transparent with interpretability and explainability, rule-based approaches become complicated when the number of rules is huge, thus spending more time and cost for maintenance, update, and retrieval. As being well-known in the medical domain, physical robots have become more collaborative with humans to be skillful and sophisticated in some advanced surgical tasks, especially when they have been trained with explainable AI to be capable of understanding and learning human guidance, consequently enhancing the trust from patients. Some robotic-based surgical procedures are gynaecologic surgery, head surgery, neck surgery, and prostate surgery.

### D. WIRELESS COMMUNICATION NETWORKS

Many innovative communication technologies have been introduced in the last decade due to the explosion of IoT technologies with edge and mobile devices along with the diversity of applications and services, especially in the healthcare and medical domains [47]. Different communication platforms and technologies have distinct features which have resolved many challenging issues and further enhanced the overall performance of healthcare and medical systems, accordingly reducing time and cost for patients, hospitals, and other healthcare institutions.

*Cellular network*: As so-called mobile network, a cell network is a radio network distributed over land areas called cells. Each cell is served by at least one fixed-location transceiver, a.k.a., cell site or base station. In this network, each cell uses a different set of frequencies from neighboring cells, to avoid interference and provide guaranteed bandwidth within each cell. Besides owning several advantages, such as flexible to use the features and functions of all public and private networks, increasing capacity, less power consumption, and other signals interference reduction, some cellular networks can join different neighboring cells to provide better radio coverage over a wider geographic region. Remarkably, cellular networks can enable a huge number of portable receivers to communicate with each other and other fixed transceivers via base stations [48]. Currently, 5G as the 5th generation mobile network can deliver higher multi-Gbps peak data speeds, ultra low latency, more reliability, massive network capacity, increased availability, and a more uniform user experience to more devices and more users, thus promoting more high-quality mobile-aided healthcare and medical services and applications [49]–[51].

*Wi-Fi*: is a wireless networking technology that allows diverse devices, such as computers (desktops, laptop, and tablets), mobile devices (smart phones, smart watches, and wearables) and others (printers, TVs, projectors, and video cameras) to connect the Internet. A Wi-Fi network is simply an internet connection that's shared with multiple devices in a home or business over a wireless router and many optional wireless access points. The router is connected directly to an internet modem and acts as a hub to broadcast the internet signal to all your Wi-Fi enabled devices, thus providing flexibility to stay connected to the internet as long as the devices inside the network coverage area [52]. Wi-Fi networks use radio technologies to transmit and receive data at high speed with different Institute of Electrical and Electronics Engineer (IEEE) 802.11 standards: IEEE 802.11b, IEEE 802.11a, IEEE 802.11g, IEEE 802.11n, and IEEE 802.11ac. As the latest generation of the Wi-Fi standard, Wi-Fi6E is an extension of Wi-Fi 6 (802.11ax) wireless standard into the 6-GHz radio-frequency band, with the important upgrades in speed and security to accelerate health informatics processing and secure medical data and patient information [53].

*ZigBee*: As based on the IEEE 802.15.3 standard for wireless personal-area networks (WPAN) and developed by the ZigBee Alliance, the ZigBee technology is designed to provide short-distance wireless solution [54] in which running wires to transfer data infeasible or cost prohibitive. This technology has many advantages, including easy setup, low power consumption, and simple implementation and integration into other devices. In [55], ZigBee was applied in a wireless sensor network and integrated into various medical devices (such as ECG and pulse oximeter) and pulse to satisfy multiple healthcare and medical requirements of bandwidth, power, security, mobility, and scalability simultaneously. To overcome the problem of signal interference in the presence of Wi-Fi network, the work [56] developed a control algorithm for ZigBee via wireless body area networks (WBANs) which were widely used for health telemonitoring systems.

*Bluetooth Low Energy*: This radio technology is designed for very low power operation. Bluetooth LE can facilitate developers a great flexibility to build products that meet the unique connectivity requirements of market with the data transmission over 40 channels in the frequency 2.4 GHz unlicensed industrial, scientific, and medical (ISM) frequency band. Especially, this technology can support multiple communication topologies, from point-to-point to broadcast and mesh, thus enabling it to assist the establishment of reliable, large-scale device networks. In addition to the support of different data transports (e.g., asynchronous connection-oriented, isochronous connection-oriented, asynchronous connectionless, synchronous connectionless, isochronous connectionless), it is now widely used as a device positioning technology to address the urgent demand of highly-accurate indoor location-based health monitoring services [57]. Comparing with traditional wired and wireless communication technologies, Bluetooth LE is beneficial for multiple goals of a wearable sensor-based healthcare systems, such as lower product cost, long battery life, and smaller system circuit while maintain high performance accuracy [58].

*6LoWPAN*: This communication technology stands for IPv6 over low-power wireless personal area network with low-power wireless modules [59]. 6LoWPAN's specification contains packet compression and other optimization scheme to allow IPv6 packets to be transmitted efficiently on a network with limited power resource and high reliability. Diverse low-power wireless networks have been introduced and implemented before 6LoWPAN, but it is recognized as one of the highly recommended protocols to strongly support IoT. This is because the IPv6-based 6LoWPAN communication allows once closed low-power wireless networks to interface with the Internet and implement more advanced intelligent services in the healthcare and medical domains [60].

### E. CLOUD AND EDGE COMPUTING

Many traditional healthcare and medical systems have faced a big problem of massive unstructured, diverse, and exponential-growing data collected from different sources, thus arising much more challenges to store and process data effectively and securely. To this end, the advanced techniques and high capacities of cloud computing allow analyzing healthcare and medical big data [61]. Due to the importance and the imperativeness of data security and privacy, the encryption and cyberattack prevention characteristics of cloud platforms are highly required. Furthermore, a general cloud architecture may decrease data redundancy and therefore should execute an efficient data curation for storage optimization in the cloud. This advancement allows the centralized database implementation and the execution of AI in smart healthcare systems. Indeed, the diverse data collected and gathered by IoT devices and stored on the cloud, that is useful to extract meaningful information and perform comprehensive data analysis. The cloud alleviates the IoT computing subsystems by executing heavy workloads at centralized systems which are usually equipped by powerful computing resources [62]. As being flexible, cloud platforms provide various computing infrastructures of storage and computation to support diverse demands of customer: infrastructure as a service, platform as a service, and software as a service. In the healthcare and medical applications, the two key scenarios of these services are big data management and data processing. As the entirety of vital characteristics of big data for precise healthcare and medical diagnosis services and applications, ongoing studies have focused on categorizing data types, generated by a variety of IoT frameworks, in a sorted manner that may be useful for deeper analysis later. The powerful computation of cloud platforms enables the deployment of complex algorithms besides extending the battery life of IoT devices and maintaining low latency with optimized computational offloading schemes [63]. There are three types of clouds, including public, private, and hybrid, but the public clouds are not appropriate to healthcare and medical applications as recommended by Health Insurance Portability and Accountability Act (HIPAA). The private clouds and hybrid clouds are much deployed for remote healthcare services thanks to their essential privacy and security besides being dynamic and adaptive with different requirement levels of security and performance (e.g., preventing clouds from cyberattacks while obtaining good performance with high computing capacity of public clouds) [64].

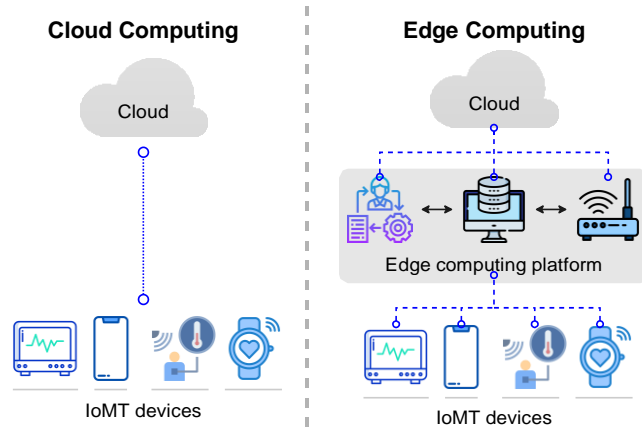

**FIGURE 2.** Cloud computing vs edge computing for IoMT-based healthcare and medical applications.

Internet of Medical Things (IoMT) devices collect an enormous amount of healthcare and medical data, which takes time and computation resources to process in the centralized systems using cloud computing. Unfortunately, not all data are useable and contain meaningful information, thus inducing inefficiency [65]. Edge computing, by sending only the most relevant and meaningful information to the cloud, can vastly reduce the workload of data centers and cloud computing resources besides enhancing operation efficiency and using lower network traffic and connectivity cost if compared with cloud computing [66], [67]. In edge computing, data preprocessing has demonstrated a vital role in catching and detecting outliers or even any abnormality of IoMT data before forwarding to the cloud. In [68], edge computing was studied for IoMT systems, in which the patient's data collected by IoMT devices were transferred to the edge layer for real-time disease diagnosis with DL. In particular, the heathcare and medical data, such as ECG, heart rate, blood pressure, glucose level, pulse rate, and chlesterol level, were gathered and sent to edge computing for pre-processing with noise removal and learning pattern with deep neural networks (DNNs). Later, the diagnosis report derived from deep models at edges were transmitted to the cloud server for hospital database, healthcare institutions, and patients as depicted in Fig. 2. In [69], edge computing was exploited for IoMT systems to achieve small-scale real-time computing and high storage capability while ensuring low latency and optimal IoMT resources utilization. Fog computing, a computing layer between the cloud and the edge, plays the role of data examination, i.e., to decide whether the edge-collected data is relevant or worth for continuously sending to the cloud, thus resulting in traffic reduction. In [70], an edge-fog computing was applied to detect epileptic seizure using lossless electroencephalogram (EEG) data in IoMT systems. The proposed approach can reduce the amount of healthcare data from edges to the fog gateway with k-means clustering and Huffman encoding and identify the epileptic seizure condition with the Naïve Bayes algorithm. As a result, the approach significantly reduced the amount of EEG data to the cloud while presenting high detection performance as well as the cloud computing-based method.

### F. IMMERSIVE TECHNOLOGY

Immersive technology, a term that refers to the technologies for reality extension using the neuroscience of the human brain, aims to create distinct experiences by merging the physical world with a digital or simulated reality. Besides augmented reality (AR) and virtual reality (VR) as two primary types of immersive technologies [71], it includes extended reality (XR), mixed reality (MR), holography, telepresence, digital twins, and first-person view (FPV) drone flight [72]. The above-mentioned immersive technologies can be briefly featured as follows:

- VR: a technology that allows the creation of a fully immersive digital environment, in which the physical or real-world environment is entirely suppressed.
- AR: a technology that enables the superposition of digital elements into the real-world environment, i.e., the composite view of physical and digital elements can be observed.
- MR: a technology that not only allows the superposition of digital elements being viewed in the real-world environment but also establishes their interaction.
- XR: an umbrella term that encompasses any types of technology that alters reality (e.g., adding digital/virtual elements to the physical environment at any level, thus nearly erasing the line between the digital world and the physical world). Indeed, XR includes AR, VR, MR, and other technologies.
- Holography: or hologram is a technology to product three-dimension image of an object with highest resolution of any imaging for device-free watching.
- Telepresence: a novel form of robotic remote control, in which a human operator can observe, feel, interact, and collaborate an object from distance.
- Digital twins: a virtual replication of real-world project, which is require for connecting the physical object for real-time data collection, processing, and response.
- FPV drone flight: a technology uses unmanned aerial vehicle (UAV) with a camera for transferring high-quality videos to goggles, headsets, and mobile devices or another screen, thus allowing users to enjoy environmental experience in the first-person view.
- Haptics: refers to a technology that uses tactile (touch) sensation to interact with the computer applications in order to uplift user experience.

In general, most of the immersive technologies are widely used in healthcare and medical training and education [73],

where there are several complicated, multi-dimensional aspects for observation, analysis, and communication. Basically, any content that requires face-to-face simulation and interaction will be suitable and highly recommended for this platform. Some examples of immersive technologies for healthcare and medical training are briefly given as follows:

- Operating machinery: Due to expensiveness and high demand, training and education with medical machinery is costly for students, trainees, and engineers, consequently increasing the diagnosis fee of patients. To this end, creating a digital twin of medical modules, devices, and even a whole diagnosis system is a promising solution, offering a safe and low-cost training environment.
- Clinical deterioration in patients: VR or immersive web can be applied to rebuild some specific scenarios, in which patients are clinically deteriorating over time. Patients can describe symptoms with clinical data and ask a treatment plan based on the diagnosis decisions [74].
- Other use cases include learning procedures, performing surgery, replicating emergency scenarios, pain management, and physical therapy.

## III. ENABLING TECHNOLOGIES OF THE METAVERSE FOR HEALTHCARE

This section presents a detailed discussion of the enabling technologies of the Metaverse for healthcare, which includes extended reality, blockchain, artificial intelligence, IoT, 5G and beyond, digital twin, big data, quantum computing, human-computer interaction, computer vision, edge computing, and 3D modelling. The illustration of the above-mentioned enabling technologies of the Metaverse for healthcare is depicted in Fig. 3.

### A. EXTENDED REALITY

Extended reality includes technologies such as augmented reality (AR), virtual reality (VR), and mixed reality (MR), aided by artificial intelligence, computer vision, and connected devices like mobile phones, wearables, and head-mounted displays [75]. By incorporating voice, gestures, motion tracking, vision, and haptics, this new technology is transforming the way services are delivered, improving the quality in various sectors. Traditionally, people thought that XR would only benefit the entertainment industry. It was expected that an immersive experience would only enhance a user's experience of a video game or movie. However, XR usage has far surpassed these expectations. It is being used in an increasing number of industries, from healthcare to manufacturing [76]. In the Metaverse, XR will realize its full potential and continue to evolve. VR and AR technologies are combined to create a sense of virtual presence in the Metaverse. The Metaverse provides virtual and immersive experiences, allowing users to move from one environment to another and perform tasks similar to those in the real world in interconnected virtual worlds. The COVID-19 pandemic has accelerated the global need for XR in healthcare [77]. The demand for virtual telemedicine and diagnosis has helped the rise in popularity of these XR devices.

The Metaverse with XR can help revolutionize healthcare. Medical students can learn more in real-life situations than in a regular classroom. But even the smallest error in the real world might have grave consequences for an individual's life. In this scenario, XR applications in the Metaverse will enable medical students to practice their abilities in a virtual 3D environment that is as realistic as the real world [78]. This interactive virtual 3D environment in the Metaverse will also help surgeons to enhance their skill level [79]. The Metaverse, with the help of XR, enables surgeons to visualize organs, tumors, X-rays, and ultrasounds in real-time and from numerous angles without losing focus on the patient while performing the operations. The Metaverse's 3D representations of a patient's body increase efficiency and accelerate the treatment [80]. In situations where an ordinary person is required to perform resuscitation, doctors can guide to perform resuscitation in the Metaverse based on the stimulated environment using XR equipment [81]. Through the Metaverse, XR and remote-controlled devices can be used to give the patient physical therapy, speech therapy, and mental health therapy. Individual health records are well protected and easy to access in the Metaverse, allowing doctors to view these records using XR devices and to conduct virtual consultations with patients to make better medication recommendations.

Although XR enabled Metaverse aid in providing better healthcare from a variety of locations, the implementation faces profound challenges. A data breach is possible in the Metaverse, as XR-related solutions have access to so much sensitive patient information, any manipulation or loss of this information will harm patient healthcare and privacy, which may also affect the doctor's reputation. Development and implementation of XR technologies that aid in the provision of healthcare in the Metaverse can be exceedingly costly. This makes the patient-doctor consultation more difficult, as not every patient can afford and utilize this innovative technology.

### B. BLOCKCHAIN

The foundation of blockchain emerged in a 2008 white paper written by Satoshi Nakamoto [82]. A blockchain is a digital database of transactions that is duplicated and dispersed over the entire blockchain network. Each block in the chain contains several transactions, and whenever a new transaction occurs on the blockchain, a record of it is added to all participant's ledgers [83], [84]. Multiple parties administer blockchain, hence it is also referred to as distributed ledger technology (DLT). A blockchain is a type of DLT that records transactions using an immutable cryptographic signature known as a hash. This implies that if a single block in a chain is altered, it would be immediately transparent that it had been altered. If hackers wanted to break into a blockchain system, they would have to change each block in the chain on all of the copies of the chain that are spread out. Non-fungible

VOLUME 4, 2016 9

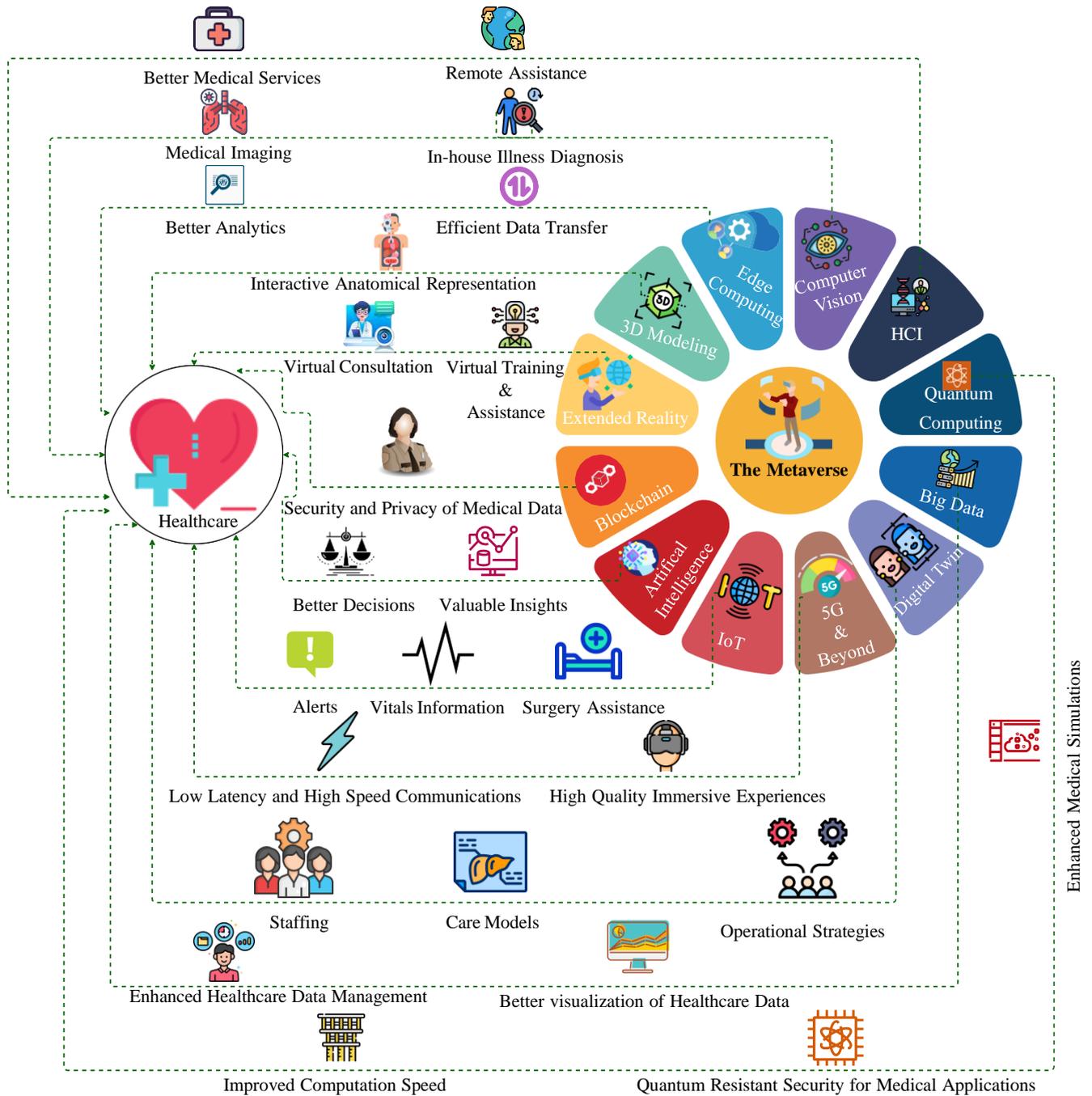

**FIGURE 3.** Enabling Technologies of the Metaverse for Healthcare

tokens and cryptocurrencies on the blockchain facilitate the creation, ownership, and use of decentralized digital assets [85]. The concept of the Metaverse would be incomplete without blockchain, as centralized data storage has several problems with security, privacy, and data transparency. The blockchain will make the Metaverse a decentralized digital source that is capable of operating on any platform and on a global scale. The blockchain-based Metaverse lets people access any digital space without the involvement of any central institution. Authentic and high-quality data collecting will be made easier with the help of blockchain technology in the Metaverse [86]. The immutability and transparency features of the blockchain will ensure that the Metaverse's enormous data is stored and managed securely [87].

The adaptation of the Metaverse will enhance the availability of doctors for patients. In the Metaverse, patients will have more options and a more conducive environment for communicating with doctors [88]. This decreases the time required to obtain a diagnosis and consultation. The blockchain-enabled Metaverse will deliver precise patient data to doctors, allowing them to make more accurate decisions. The blockchain will secure such sensitive information from being modified or tampered with by attackers [89].

There are also some constraints to the blockchain-enabled the Metaverse for healthcare. The adoption of this technology has challenges due to its huge cost and resource requirements. This technology is completely unrestricted and unregulated, which puts patients at risk. The technology's complexity will make it difficult for end users or patients to adapt to it. Due to the necessity of storing all data at every node linked to the chain, the blockchain-enabled Metaverse might be extremely slow [87]. Small hospital chains cannot use this technology because of its excessive energy usage and complexity.

### C. ARTIFICIAL INTELLIGENCE

Artificial intelligence (AI), also known as machine intelligence, focuses on the development and management of technology that can autonomously learn to make decisions and carry out tasks on behalf of humans [90]. AI is a collection of technologies that incorporates any software or hardware component that facilitates machine learning, computer vision, natural language understanding (NLU), and natural language processing. AI will help strengthen the Metaverse infrastructure, enhancing the 3D immersive experience, and boosting the virtual worlds' built-in services. AI technology will also help in improving the quality of services and the Metaverse ecosystem.

The health industry has recently begun utilizing revolutionary techniques such as XR and big data combined with AI in software and hardware to increase the efficacy of medical devices, reduce the cost of health services, enhance healthcare operations, and broaden access to medical care [91]. The Metaverse enables immersive learning, understanding, and sharing of patient's health issues and medical data with their doctors. AI is being utilized to analyze/diagnose patients' health data. The Metaverse, with the help of AI, facilitates doctors with high-quality 3D images and scans of the patients required for intervention. AI can help in providing crucial insights to the doctors that can help in prioritizing critical patients, minimize potential errors in analyzing electronic health records, and create more accurate diagnoses [92]. Huge volumes of health data and medical records make it difficult for doctors to stay focused on the latest medical advancements and provide quality, patient-centered treatment. Electronic health records and biomedical data collected by medical units and medical professionals can be rapidly analyzed by AI algorithms in the Metaverse to offer doctors immediate and trustworthy recommendations [93]. The collaboration of the Metaverse and AI can also help in drug discovery, disease forecast, and emergency response.

An AI-enabled Metaverse might pose significant risks to patient privacy and ethical issues, and even cause medical errors, which can mislead the doctors in providing treatment to the patients, although it can open up new insights and expedite interactions with healthcare data for both patients and clinicians. The lack of justification for results will also pose a significant challenge to adopting the AI-enabled the Metaverse [94], [95].

### D. INTERNET OF THINGS

The phrase "Internet of Things" (IoT) refers to the billions of devices that are currently connected to the Internet and exchanging data [96]. Due to the introduction of low-cost computer chips and the broad availability of wireless networks, everything from smartphones to intergalactic operations can now be connected to the IoT. Due to the incorporation of sensors and the capacity to communicate with one another, these devices can share real-time data without the need for a human supervisor [97]. The IoT, with the help of the Metaverse will bring the digital and physical worlds closer together, resulting in a more intelligent and responsive environment.

In healthcare, remote patient monitoring is the most widespread application of the IoT enabled Metaverse. Patients who are not physically present in a hospital or clinic can have their vital health metrics such as heart rate, blood pressure, temperature, and more automatically collected by IoT devices and displayed in the Metaverse's 3D environment. This eliminates the requirement for people to travel to see their healthcare professionals or to collect their vital health data [2], [98]. By deploying small Internet-connected nanobots inside the human body and monitoring them in a virtual 3D environment enabled by the Metaverse, surgeons will be able to perform complex procedures that would be difficult to perform with human hands alone. At the same time, robotic surgeries done using small IoT devices or nanobots can reduce the size of the incisions needed for surgery. This makes the process less invasive and allows patients to heal quickly. The collaboration of the Metaverse and IoT can also help in chronic disease management, improving the sleep cycle, medication refills, and alerts in the time of health emergencies.

The issues related to data security and privacy are very crucial in the field of healthcare [99]. Data security and privacy issues will be a concern of IoT-enabled Metaverse as IoT-enabled devices don't comply with data protocols and standards in real-time. The integration of IoMT devices with Metaverse will also be a challenging task. Due to the non-uniformity of data and communication protocols of the IoT-enabled Metaverse, it will be difficult to aggregate data for vital insights and analysis. The cost of these technologies will also be another challenge for adapting IoT-enabled the Metaverse into the field of healthcare.

### E. 5G & BEYOND

The term "5G and beyond" refers to the fifth generation and beyond of wireless technology. It outperforms 4G LTE networks in terms of both speed and latency, and it has a higher capacity. In contrast to the maximum speed of 4G technology, which is only 1 Gbps, the maximum speed of 5G technology can exceed 20 Gbps. Additionally, 5G

VOLUME 4, 2016    11

delivers reduced latency, which has the potential to enhance the performance of commercial applications and other digital experiences [100], [101]. Its higher speed and greater connectivity will benefit the productiveness of businesses. Networks that go beyond 5G are important for all sectors from developing smart healthcare to smart cities. The 5G and beyond technologies will bring the Metaverse closer to realizing its true potential. The creation of rich media content using immersive technologies like AR, VR, and the IoT, the Metaverse requires low latency and high bandwidth connectivity which is provided by 5G and beyond technologies. As a result, 5G and beyond technologies will have a big impact on the Metaverse's ability to produce new kinds of content with greater ease [102].

The healthcare sector can be benefited from advancements facilitated by the Metaverse. Patients and physicians will be benefited from the high-quality immersive experience provided by the Metaverse with the help of 5G and beyond technologies. The high speed and low latency communication provided by the 5G and beyond technologies to the Metaverse will enhance many aspects of the healthcare sector, which includes virtual wellness, mental health, and complex surgical procedures [103]. The 5G and beyond technologies enables the Metaverse to improvise the opportunities for uninterrupted immersive medical education and training of the medical students. Using current wireless networks, remote surgery in the 3D environment provided by the Metaverse is perilous because of the latency issues because of poor speeds. Patients are at risk if this process is prolonged. With its ultra-low latency, 5G and beyond networks promises to reduce the lag period between a remote surgeon's input and real-time action, which eliminates risk from remote surgeries [104].

Although 5G and beyond enable the Metaverse to provide faster and latency-free communication in the field of healthcare, the cost will also increase due to the adaption of the 5G and beyond enabled for the Metaverse. In addition, the flow of data will also increase, which requires huge computational and storage resources. The scarcity of 5G-enabled devices will also have an impact on the Metaverse adaptation in the fields of healthcare.

### F. DIGITAL TWIN

A digital twin is a virtual representation that acts as the digital counterpart of a physical object or process in real-time [105]. Digital twins were coined in the year 1991 in the book Mirror Worlds by David Gelernter [106]. The concept was also mentioned by Michael Grieves in the year 2002 [107]. The first practical definition of a digital twin was proposed by NASA in 2010 to enhance the physical-model simulation of spacecraft [108]. Digital twins are the outcome of the continuous advancement of product design and engineering processes. A digital twin is a digital representation of an object, process, or service that exists in the actual world. A digital twin can be a digital copy of a physical thing, such as machinery, medical devices, or even larger objects, such as skyscrapers or even entire towns. The Metaverse technology, on the other hand, represents a virtual world in which everything and everyone interacts similarly to how they do in the actual world. Digital twins are building blocks of the Metaverse as they create a digital replica of every object in the Metaverse [109].

Operational strategies, staffing, and care models can be examined by establishing a digital twin of the whole hospital in the Metaverse to assess the requirements. In scenarios like lack of beds, the transmission of pathogens, the scheduling of doctors, or the availability of operating rooms, these virtual models in the Metaverse can help. Patients' treatment, costs, and performance of the staff can be improved by using digital twin-enabled Metaverse. Healthcare is a very complicated and sensitive environment, so this is critical for making strategic decisions. The hospital can be fully virtualized in the Metaverse using digital twins to offer a risk-free environment. The digital twins-enabled Metaverse will also help in creating personalized artificial organs [110]. Digital twin-enabled Metaverse can also help to perform brain and heart surgeons in virtual simulations of surgical procedures before executing complex real-world surgeries [103].

The challenge of creating a virtual replica of every real-world object requires the creation of a large number of digital twins along with huge computational resources. The most difficult challenge is to create an organ or live creature's virtual replica in the Metaverse. Making a replica of a person with all of their organs and systems working together in sync in real-time is a big challenge for the digital twin-enabled Metaverse in healthcare.

### G. BIG DATA

Big data is defined as the "Information asset characterized by such a High Volume, Velocity and Variety to require specific Technology and Analytical Methods for its transformation into Value." [111]. Analyzing such enormous amount of data collected from various sources at high velocities enable generating value out of the same [112]. A detailed review on the various underlying concepts of big data have been studied by the authors in [113]. However, big data analysis and analytics often poses various challenges as explained in the works by [114] [115] [116]. The use cases of big data spans across various fields such as healthcare, industry, education, agriculture and so on, thus providing better sentiment and behavioral analytics [117], predictive support [118] and fraud detection [119] [120]. The Metaverse, however, enables a persistent, real-time and interoperable environment with which people can interact and explore in a virtual world [121].

Big data was considered as the next big thing in computing; the Metaverse is now, the next big thing in big data. Big data is a key technology in the Metaverse, that is growing rapidly and is gaining rapid momentum with the intervention of the Metaverse itself. The Metaverse includes data in structured, semi-structured and unstructured formats, that is difficult to be handled using traditional data analytics tools. However, integrating big data analytics with the Metaverse will aid organizations to build a single point cloud-based platform through which internal and external data can be collected and analyzed for generating valuable insights. This characteristic is indeed important in the Metaverse, as real-time decisions need to be taken and the facility should enable foreseeing the future, thus making accurate predictions. As valuable insights can be obtained with various analysis techniques, handling enormous data generated with the Metaverse and its underlying mechanisms require efficient big data tools. This becomes increasingly important in healthcare scenario, as digital simulations of the patient can be created with the help of which interesting patterns can be unraveled [122]. The Metaverse in healthcare can provide accurate predictions along with the likelihood of the effect of a particular medicine/ treatment on the patient. Integrating big data analytics tools in such scenarios would help in understanding how exactly the target group will respond to the current or future decisions.

Integrating big data technologies with the Metaverse has already caught the attention of various researchers and few studies were also published on its applications in various domains. One such study was presented by the authors in [123] that details about the various attributes of the Metaverse in education and put forwards the Edu-Metaverse. Similarly, the authors in [124] have explained how the Metaverse and big data are utilized for business operations in Indonesia, along with the underlying challenges and opportunities. Even though the research in the big data enabled-Metaverse is still in its nascent stage, it becomes extremely essential, as the enormous amount of real-time data in a variety of formats that could be generated by the Metaverse can be handled efficiently with the integration of various big data technologies.

### H. QUANTUM COMPUTING

Quantum computing is a form of computation whose activities can exploit quantum mechanical phenomena such as superposition, interference, and entanglement. Quantum computing devices are referred to as quantum computers [125]. Quantum computing is capable of tackling various computational tasks. Quantum computing helps in critical decisions significantly quicker than classical computing. So, quantum computing is gaining popularity in a variety of fields, including banking and pharmaceuticals [126]–[128]. The Metaverse, on the other hand, intends to integrate many sectors into a single ecosystem to provide a virtual environment that closely resembles its real-world equivalent. The objective is to build an interface that mixes social networking, gaming, and simulation to create a realm that mimics reality. In particular, digital healthcare has emerged as a result of the pandemic's impact on medical consultation. Realistic interactions between patients and physicians will use the Metaverse realism and high-level visuals. This feature of the Metaverse will provide frequent engagement between patients and physicians in this digital environment. Enormous computation power is needed to construct these virtual worlds. The Metaverse helps create an interactive virtual healthcare community that resembles reality in appearance and experience.

The quantum computing-enabled Metaverse will make it possible to overcome the enormous challenges related to security, large computation capacities, and cyber attacks. Through the development of quantum-enabled security applications, quantum computing will provide the Metaverse with the required security. All activities and transactions in the field of healthcare will require quantum-resistant security as the Metaverse expands. The enormous computational capabilities of the quantum computing-enabled Metaverse will boost the efficiency of healthcare applications. The healthcare applications in the Metaverse make extensive use of computation and simulation. Hence, quantum computing improves computation and the overall experience of doctors and patients. Quantum randomness enables developers to ensure that the regulations put in place prevent inhabitants and algorithms from manipulating the healthcare applications in the Metaverse [129].

Quantum computing has not yet attained total dominance, and these machines may not appear for several years. Quantum computing employs qubits, whereas current Metaverse-enabling technologies and systems are based on binary systems. This significantly complicates the integration of quantum computing with the Metaverse. Quantum computing necessitates a tremendous amount of energy, a problem that must be resolved to realize the full potential of its integration with the Metaverse. If quantum supremacy were to fall into the wrong hands, all cryptography and security mechanisms of the Metaverse would collapse.

### I. HUMAN COMPUTER INTERACTION

Human-computer interaction (HCI) is a multidisciplinary topic of research that focuses on the design of computer technology and, specifically, the interaction of humans and computers. Voice, gesture, visual, and brain signal interaction have replaced textual or display-based control as the dominant paradigm in HCI [130]. HCI, VR, AR, and the future of content creation and collaboration technologies will enable the creation of the Metaverse. The visual interactions in the Metaverse will be carried out by HCI technology which is called wearable consumer head-mounted displays (HMD). These HMDs will play a crucial role in the communication between the users and surroundings in the Metaverse. These HCI devices will enable the users' senses in the Metaverse. Another HCI technology called haptic wearable device will give the user touch, smell, and taste experiences in the Metaverse [131]. These devices will also enable the users to perform tasks with help of robots from remote locations using the realistic environment provided by the Metaverse.

Technologies like holographic building, holographic emulation, XR integration, and XR inter-connectivity will be included in the HCI-enabled Metaverse. In several ways, healthcare will benefit from the HCI-enabled Metaverse. It will enable medical education for the students, popularize research findings, improves consultation, and provides accurate graded diagnosis, and better treatment by improving interactivity and immersion provided by the HCI-enabled Metaverse. The users will receive better medical services such as illness prevention, telehealthcare, physical examination, disease diagnosis and treatment, rehabilitation, chronic disease management, in-home care, and first aid with the help of the HCI-enabled Metaverse. The procedures can be performed by surgeons from a remote location with the help of robots in the realistic environment provided by the Metaverse.

Although HCI technologies help the Metaverse in various ways, these technologies still pose numerous challenges. Modern HCI devices like head-mounted displays and haptic wearable devices are still immature; they require a significant amount of time to allow the Metaverse to realize its full potential. These HCI devices do not comply with any standards or norms, making their integration into the Metaverse challenging. The affordability of these devices will be an additional obstacle, making it harder for people to adapt to them.

### J. COMPUTER VISION

Computer vision is the study of how computers visualize and interpret digital images and videos. Computer vision encompasses all activities done by biological vision systems, including perceiving a visual signal, interpreting what is being seen, and extracting complicated information in a form accessible by other processes. Computer vision uses sensors and machine learning algorithms to simulate and automate essential parts of human visual systems [132]. Autonomous vehicles, facial recognition, image search, and object recognition are some of the popular applications of computer vision. In the Metaverse, computer vision helps in recreating the user's physical experience in the 3D virtual world using technologies including object identification, plane detection, facial recognition, and movement tracking. Computer vision will help in determining the user's position and direction in the Metaverse. In the Metaverse, users will be portrayed as avatars with help of computer vision. The interactive computer vision system will continuously track the location and movement of the user in the Metaverse. Users are required to engage with their surroundings in the Metaverse. Consequently, image processing with the help of computer vision is essential to the development of more immersive Metaverse experiences. Computer vision enables a seamless connection between the Metaverse and the physical world in real-time. The avatar created must be collaborative with the users in all situations, or the user will be frustrated if they are not in control of the surroundings. Computer vision ensures the Metaverse accurately portrays the 3D virtual environment by providing immersive visual experiences to the users.

The computer vision-enabled Metaverse will aid in medical imaging with the use of IoMT devices. It can aid in tracking and visualizing the tumor and cancer cells in a 3D environment. It can aid medical practitioners in their training by offering diverse scenarios. The computer vision-enabled Metaverse can aid in combating pandemics like Covid-19 by enabling in-house illness diagnosis and reducing the global spread of the disease. It also helps the doctors in remote patient monitoring and makes decisions based on the vitals information provided by computer vision-enabled Metaverse.

The integration of computer vision with the Metaverse also poses a few challenges. The techniques used by the computer vision-enabled Metaverse may not accurately project the surrounding which may result in distorted surroundings in the Metaverse. Vision abnormalities such as farsightedness and astigmatism, insufficient eye focusing or eye coordination abilities, and age-related eye changes such as presbyopia can all contribute to the development of visual symptoms when using a distorted imaging display. Computer vision enabled Metaverse can provide great enhancements in surgery and therapy of some diseases but any smallest of mistakes will lead to the grave consequences.

### K. EDGE COMPUTING

Edge computing is a distributed computing paradigm that moves computation and data storage closer to the data sources. Edge computing will improve response times and conserves bandwidth [133]. Transferring all of the device-generated data to a centralized data center or the cloud raises issues with bandwidth and latency. Edge computing provides a more efficient option, as data is processed and analyzed closer to its origin. Reduced latency results from the fact that data does not travel over a network to a cloud or data center to be processed [134]. Edge computing, including mobile edge computing on 5G and beyond networks, provides faster and more thorough data processing, allowing for deeper insights, quicker reaction times, and enhanced user experiences.

The physical world's objects will be incorporated into a new digitally-created virtual environment in the Metaverse. This entails collecting enormous quantities of data from several devices. The created data must be sent to a centralized location or the cloud for storage. The number of users and devices, together with the rate of data transmission, will expand significantly. Any delay in data flow or storage will result in catastrophic failure of the Metaverse applications. Edge computing will enhance the transmission rate of the Metaverse and decrease bandwidth use [135]. This assures the emergence of the Metaverse.

The Metaverse with edge computing capabilities can maximize transmission rate and also store a limited quantity of data. In the Metaverse, edge computing brings data processing, analytics, and storage closer to the data generation source. This technique can aid health applications in the Metaverse to optimize the gathering, storage, and analysis of data by leveraging edge computing capabilities. Wearables like trackers and smart watches will be able to efficiently transfer data and provide doctors with an up-to-date status of essential patient vitals such as heart rate and blood pressure, therefore alerting medical professionals to potential problems before they arise. Health monitoring systems can facilitate remote patient care by gathering patient data and activating actions in real time depending on the results. Real-time uninterrupted collecting of data associated with imaging models can aid in the detection of possible issues in X-rays, hence prioritizing those images for radiologist or physician examination and providing necessary treatment from remote locations.

Edge computing will boost the Metaverse's data retrieval capabilities on a real-time basis. It also necessitates the addition of a large number of additional edge devices to the existing network which is challenging. It also posts several challenges related to security, data loss, heterogeneity among devices, and cost [136].

### L. 3D MODELING

3D modeling is the process of producing a mathematical coordinate-based representation of any three-dimensional surface of an item using specific 3D modeling techniques. As a result of advancements in image processing, computer-aided design, and modeling techniques, it is now feasible to create extremely realistic and trustworthy 3D models [137]. Multiple sectors, including cinema, animation, and gaming, as well as interior design and architecture, utilize 3D modeling. The fulfillment of the Metaverse is reliant on the quality of 3D replication of real-world objects in the virtual world. The Metaverse objects represented in the 3D range from people to cities to jungles. All the Metaverse-related applications will rely on 3D modeling.

In the healthcare industry, interactive anatomical representations of high-quality imaging may be created using 3D modelling-enabled Metaverse. Patients can view medical assistance products in 3D, allowing them to purchase the appropriate equipment. The Metaverse's 3D modeling capabilities can assist doctors in gaining a deeper understanding of a patient's illness. These findings will aid physicians in providing better care to patients and performing treatments more precisely. The 3D modeling capabilities of the Metaverse allow medical researchers to create artificial organs and evaluate their performance in various simulated environments.

Although 3D modeling assists the Metaverse in realizing its full potential, it presents several obstacles throughout this transition. This transit requires a tremendous amount of man and machine hours. The Metaverse is an ever-growing virtual environment that has no bounds for content and creation, which is a never-ending challenge for 3D modeling techniques. Objects modeled for the Metaverse will be dynamic, necessitating that the accompanying 3D models are dynamic as well.

## IV. POTENTIAL APPLICATION OF THE METAVERSE FOR HEALTHCARE

The Metaverse in medicine significantly helps in providing comprehensive healthcare when compared to the "handicraft workshop model" where the diagnosis and treatment varies from doctor to doctor and hospital to hospital. In a compre-hensive healthcare scenario, decisions will be made based on the expert's suggestions and the results obtained from the various enabling technologies of the Metaverse. The Metaverse in medicine has numerous applications ranging from research, physical examination, and diagnosis to insurance. Some of the plausible implementations of the Metaverse which have potential to gain momentum in the near future includes virtual physiotherapy, virtual biopsy, virtual counselling and virtual alert response. Virtual biopsy is a non invasive technique to characterize tissues by obtaining and processing the image. Virtual physiotherapy would help the rehabilitation patients in guiding them through movement and exercises. This section discusses the applications of the Metaverse in comprehensive healthcare which is depicted in Fig.4

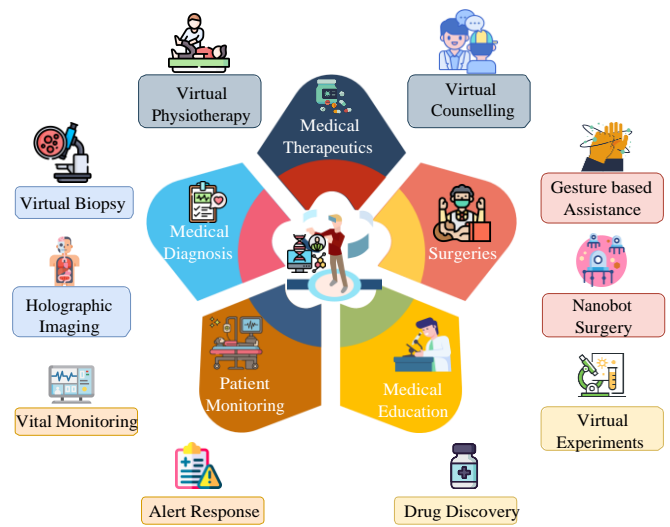

**FIGURE 4.** Potential Applications of the Metaverse in Healthcare

### A. MEDICAL DIAGNOSIS

Medical diagnosis is the process of determining the medical condition of a patient based on the symptoms. Adoption of the Metaverse in healthcare significantly helps in efficient diagnosis of the medical conditions of a patient with the help of various advanced technologies such as AR and VR enabled MIoT models, extended digital twin, blockchain, 5G and so on. An article on "expert consensus on the Metaverse in medicine" was put forwarded by [30], that explains how and why the Metaverse can be applied to different verticals in healthcare for delivering quality comprehensive healthcare for all. The Metaverse can also be considered as an enhancement of the existing medical IoT that overcomes its limitations pertaining to human-computer interaction, interconnection and integration with and between real and virtual worlds.

MIoT accessed using the AR/VR glasses can help in holographic construction, emulation, real and virtual world interaction and integration, thus simplifying the complex problems faced in a healthcare scenario. The authors in [138]

VOLUME 4, 2016  15

have provided a comprehensive study on the applications of holography in medical field. Intelligent diagnosis can be made using the Metaverse, since the experts in both the real and virtual worlds can guide the medical professionals in delivering quality healthcare and diagnosis by communicating with them and being present in any place in a virtual manner thereby having a treatment procedure that is in-line with the standards put forwarded by the medical agencies. Integration of blockchain technology with distributed ledger technology in the Metaverse will enable an efficient means of storing, exchanging and programming health-based digital assets across multiple platforms, with the help of which the data can be used for arriving at wiser and precise diagnosis of various medical conditions. Various applications of the Metaverse in ophthalmology have been researched by the authors in [139]. Also, [140] put forwarded a review on the adoption of blockchain for ophthalmologic usecases.

"BlockNet" is a secure, multi-dimensional data storage solution that improves the data reliability of digital twins using blockchain technology [141]. A "nonmutagenic multidimensional Hash Geocoding" method is employed here for efficient indexing of multidimensional data. The possible applications of BlockNet in healthcare is in the microscopic analysis. As diagnosis is the key and basic procedure in determining the type of treatment and medications to be provided, employing the Metaverse in diagnosis will greatly help in the overall quality improvement of all the other phases.

### B. PATIENT MONITORING

The convergence of telepresence, digital twinning and blockchain will reap amazing benefits of the Metaverse in healthcare, especially in terms of patient monitoring. Telepresence in medicine, also called as telemedicine, provides medical services remotely [142]. A digital twin is a virtual, digital model created using the real-world data of an object/ system/ person that helps to learn more about its real-world counterpart. These test dummies of patients can be used in case of critical situations, so that even the responses to surgeries or medicines can be known way before it is provided to the patients in real. Since medical data is the most sensitive and important, using blockchain technologies aid in storing and transferring them securely so that the data will not be tampered and is not put as risk.

Patient monitoring can be provided effectively if all these three components are made to work together in an effective manner. The Metaverse comes to the rescue by converging these technologies as a single entity. CoViD-19 has made the medical practitioners think about delivering quality healthcare even remotely by providing medical advises followed by a video call or voice call with the patient [143] [144] [145]. However, with the emergence of the Metaverse in science, healthcare sector will be benefited immensely by having the virtual world created where ever necessary and provide treatments to the needy even when they are continents apart. The AR/ VR mechanisms provide a sense of "being at" the place and this is beneficial in case of patient monitoring. This has applications not just for the interaction between the healthcare practitioners and the patients, but also between the patient and his family members. As family can play an important role in the overall improvement of the health condition of a patient [146], even if the families are at different places, a feeling of "being together" can be made possible with the help of the Metaverse. Thus, patient monitoring using the Metaverse can significantly improve the health condition of a patient by having quality interactions between patients, healthcare practitioners and family members, thereby creating a positive environment for the patient himself.

### C. MEDICAL EDUCATION

The Metaverse is a remarkable milestone in the field of medical education. IoT, blockchain, AI, AR and VR are the pioneers of the Metaverse in the field of medical education. Contribution of AI, blockchain and the Metaverse in medical healthcare was presented in [29], where the unique identifying tag in blockchain helps identify data in BC based Metaverse. The Metaverse uses AI and blockchain to create a digital virtual world that transcends the limits of the real world. These technologies allow the medical students even in their busy clinical environment to focus on the session, participate in discussion, interact in detail and involve with more enjoyment. In the conventional methods of teaching where the tutor takes the medical students to a patient and the associated medical information would be presented and discussed among the batch of students. With the boom of these digital integration and 3D technologies there is a major change in the clinical teaching where the patient in the virtual mode is carried to a batch of medical students.

In addition to surgery, doctors are using virtual reality for examining colonoscopies, providing training to the medical students, and many other applications. The common thread is the need to intuitively understand and interact with anatomy and ultimately improve patient care.

Medical students and many doctors learn body structure by using virtual reality. One can completely analyze the human body, starting from the skeleton, nervous system, muscles, and much more. This kind of learning offers unique possibilities, and the quality of future doctors is heightened.

The technology used by the physicians for the augmented reality surgeries consists of a headset with a see-through eye display that projects images of the patient's internal anatomy such as bones and other tissue based on CT scans, essentially giving the surgeons X-ray vision.

VR allows the user to create a patient-specific 360-degree reconstruction of the anatomy and demonstrate to patients the surgical plan of their operation. This approach helps you improve the understanding of treatment and ensure the trust of patients.

The COVID-19 pandemic has disrupted the education of the medical students and practitioners extensively. Due to several factors associated with the pandemic situation the number of surgeons in-house, number of cases and in-person training opportunities were all affected [147].

Mirror World (MW) is defined as a constructed copy of the real world in digital form [148]. MW is a block chain based autonomous virtual world where the information, appearance related to the application are embedded in it and reflects as if a mirror. The MW makes the real world more convenient and efficient. Examples of representative mirror worlds used in education include "digital laboratories" and "virtual educational spaces" created in various mirror worlds.

"Virtual educational spaces" is one of the example applications of MW used the field of medical education. In [149] the authors have deployed the MW simulation in medical education. The information regarding the different portals such as operation room, classroom and meeting room are video recorded and embedded in MW simulation. The respective portals helps the student community in different ways. The operation room portal helps the student community to understand and experience the surgical room environment from various dimensions. The class room portal captivates the student community with lecture sessions and the meeting room portal accommodates the maximum population for discussion.

David Baker's team at the University of Washington, which studies protein structure, has used this digital lab to have people fold protein amino acid chains. Through this game, where the protrusion structure matches well, and the player gets points and ranks up if they succeed, the protein structure is found for an AIDS (acquired immunodeficiency syndrome) treatment, and the achievement of 60,000 participants in 10 days was described in a journal publication [150]

Augmented reality may find its way into the primary care clinic, operating room, emergency room, and dental office. Doctors could use it, for example, to plan plastic surgeries and other complex operations. They also could use it to help guide them during surgeries of various kinds.

### D. SURGERIES

The Metaverse is becoming an important technology in the medical industry, especially in surgery. Surgeons are currently using tools that range from VR headsets to haptic gloves to mimic real surgical procedures; boosting preparedness and efficiency in the operating room.

Augmented Reality can help surgeons perform surgeries more conveniently by giving them hassle-free access to data. Augmented reality can provide surgeons with fast, easy, and hands-free access to patient information by mapping 3D virtual models onto the patient body. Lecturers and professors could teach the complicated surgeries in three dimension in the Metaverse.

The authors in [151] presents a AR based product design that aims to perform maxillofacial bone surgery. The AR system comprises of head mounted wearable device that captures the visual features of the human and facilitate maxillofacial bone surgery. The product design helps the practitioners to make a virtual planning by superimposing the patient details.

The authors in [75] presents how the Metaverse technologies are accelerated in spine surgery. For past two decades the surgeons had to adapt Minimally Invasive Spine Surgery (MISS) to perform spine surgery and this method had high rate of radiation exposure. Lack of navigation guidelines, and indirect visualisation are the accelerating factors for the digital transformation in spine surgery. The 3D reconstructed images during spine surgery can only be seen in 2D flat monitor. The viewing technologies that support digital transformation in spine surgery are 3D hologram with high spatial imagination.Surface topography is used for postural analysis. It is gaining attention because of its diagnostic accuracy and low radiation dose. Wearable sensors(IoMT) used to collect data from the patients and facilitate to monitor the patients. Utilization of these enabling technologies of the Metaverse in the field of healthcare systems will benefit the patients, surgeons and students.

The authors in [152] have developed a prototype of an aneurysm clipping simulation namely, Immersive Touch Aneurysm Clipping Simulator (ITACS). The prototype is designed to provide the user, a haptic VR clipping for cerebral aneurysm. The simulator has the capability to sense the wall of the aneurysm, the brain tissue, the parent artery, and also to mimic the aneurysm rupture. Also the simulator helps the neurosurgical residents have better understanding about the aneurysm anatomy of the patients. The simulator is designed in such a way that it provides fine tactile details like sensory feedback to the users.

### E. MEDICAL THERAPEUTICS & THERANOSTICS

Medical therapeutics can be regarded as the branch of medicine that deals specifically with the treatment of diseases. Digital therapeutics (DTx) deliver evidence-based therapeutic interventions, and can be considered as a class of digital medicine. Digital Therapeutics Alliance defines digital therapeutics as products that "deliver evidence-based therapeutic interventions to patients that are driven by high quality software programs to prevent, manage, or treat a medical disorder or disease". This involves the use of various digital technologies to maintain the physical and mental well being of patients. The authors in [153] gives an overview of the past trends and the future prospects of digital therapeutics. Medical theranostics, however, is a combination of the terms therapeutics and diagnostics. An interesting work on the evaluation of the term "theranostics" along with its different definitions were put forwarded by the authors in [154].

Intervention of the Metaverse in therapeutics and theranostics can bring noteworthy changes in the field of medicine with the help of the underlying technologies. The Metaverse can improve the physical well being of a patient by making them do physical exercises. As digital therapeutics avoids the use of drugs to treat a patient, it becomes extremely important in the field of healthcare. Computer Vision is one such technology that could process, analyze, visualize and interpret images and videos. The Metaverse is the future of medicine that can indeed take telemedicine to the next level by integrating the technologies such as AR, VR, ER, blockchain, AI, and computer vision. Digital twin of a patient can be created using the existing electronic health record (EHR), with the help of which a 3D simulation can be obtained. EHRs thus plays a significant role in the healthcare domain.

## V. ONGOING AND UPCOMING PROJECTS

The Metaverse in healthcare will be mainly fuelled by three channels: Blockchain, Digital Twins and Telemedicines [162]. In this section, we will discuss different ongoing and upcoming projects of the Metaverse which are carried out/announced by various renowned healthcare companies around the world [163].

### A. HEALTHIFY

HealthLand.io, known as Healthify, has started a Blockchain the Metaverse project with the objective of providing their customers health trips, mental and physical wellness, virtual clubs and meeting with heroes. Moreover, this project will provide the healthcare experts to open online gyms and the customers/patients to access online stores. The mission is to build a virtual place where the people around the globe come on a single platform for a noble cause by getting rid of their race, religion or nationality. [164].

### B. DEHEALTH

With the aim of simplifying the healthcare ecosystem, a British organization, DeHealth, has conceptualized a decentralized the Metaverse. The technology will extend virtual reality (VR), augmented reality (AR) along with mixed reality (MR). This platform will provide a virtual world for doctors and patients to work, to interact with one other as well as to earn virtual money through the selling of anonymized healthcare information. Moreover, this project will revolunize the access of healthcare services around the world through a novel web 3.0 protocol framework [165].

### C. BUMP GALAXY

Bump Galaxy is a mental healthcare the Metaverse which is prototyped on Minecraft. This project aims at addressing mental health issues of patients, with the adoption of gaming environment. This 'game world therapy' will not only help the patients to feel societal support and safety, but also will assist the patients to overcome depression, anxiety or trauma. The visualizations with deep hypnosis will develop mental resilience and well-being among the community via this bottom-up model [166].

### D. ACCUVEIN

AccuVein is a vein visualization project which will prove a game-changing technology for clinicians, neurosurgeons and patients as it will bring significant improvement in vascular access rate. This will not only increase annual savings after clinical procedures, but will also increase the success rate of first-time injection. Moreover, patients will get rid of fear of bruising after the vein access procedure, while surgeons will be able to see a gigantic and complex network of blood vessels. This project will surely enhance treatment quality and safety in time-critical healthcare applications [167].

### E. HINTVR

California-based tech startup 8Chili has developed a novel the Metaverse platform, HintVR, which aims at serving doctors, surgeons, patients, clinical staff, paramedical staff, medical trainees as well as content curators. Surgeons are not only provided advanced 3D visualtization for image-guided surgeries, but also, they are provided innovative and immersive ways for engaging patients during pre-operation and post-operation procedures. Thus, it is a promising platform for: (i) doctors to simplify consultation; (ii) surgeons to perform non-invasive surgeries efficiently; (iii) patients to understand the diagnosis, treatment options as well as testimonials; (iv) medical/paramedical staff to effectively follow each details from the experts; (v) trainees to efficiently learn with 3D immersive training; (vi) content curators to easily curate contents and monetize [168].

### F. HEALTHBLOCKS

HealthBlocks is a Dutch startup project on IoTex Blockchain platform which aims to get rid of global healthcare inequality. It has not only remote patient monitoring feature but also has the capability to assist the patient. Smart devices will have decentralized identities to create a novel ecosystem with high level of privacy and security. This Web3 the Metaverse application will allow the users around the world to earn rewards by adopting a healthy lifestyle and changing their daily unhealthy habits. This project promises to bring higher scalability and higher transaction speed with lower transaction cost [169].

### G. OTHER PROJECTS

Microsoft will soon bring its project 'Mesh' in order to redefine the medical education and training across the globe [170]. Meanwhile, Microsoft 'HoloLens 2' can be integrated with Azure Remote Rendering to develop some stand out contents viz. Holoportation, Avatars for healthcare services. In addition, healthcare workers can consult remote experts with 3D Magnetic Resonance Imaging (MRI) images to greatly enhance quality of healthcare operations [171] [172].

Intuitive Surgical is a leading company in developing minimally invasive solutions. It has come up with 'IRIS' product which adopts AR/VR technology not only to provide medical education and training, but also, to provide critical details of a clinical image to remote expert [173]. Meanwhile, Global Healthcare Academy (GHA) has partnered with 8chili with the objective to bring immersive medical training and education using the Metaverse.

TABLE 2.

| Ref. No | Enabling Technology | Contributions | Challenges | Future Directions |
|---|---|---|---|---|
| [28] | Blockchain, XAI, Teleoperation, 6G | A framework for blockchain and XAI assisted telesurgery is proposed. 6G TI channel is also used. | Real metaverse set up is required for testing the framework | Different XAI models can be compared and the optimal technique can be chosen |
| [155] | Extended reality, Mixed Reality | Eye MG Holo: An immersive 4D pedagogical tool for learning about various ophthalmologic structures is proposed. | Cost effectiveness: Approximately 3500 USD for a HoloLens 2 | Can be used for surgical simulation training |
| [156] | Extended reality, Virtual Reality, Augmented Reality | A training in lung cancer surgery using metaverse is explained. The smart operating room was set up in Seoul National University Bundang Hospital, South Korea. | Advanced imaging and other high-end equipments are required for accomplishing the task | Can be extensively used for surgical training and other health related applications |
| [88] | Augmented Reality, Virtual Reality | Cardioverse is introduced for the diagnosis and prevention of cardiovascular diseases | Legal Regulations, security and privacy, user rights | Moral and credibility aspects need to be considered |
| [157] | Machine Learning | A hybrid Structural Equation Modelling- Machine Learning approach is proposed to predict the intention of specific users to employ metaverse in healthcare education. Application of metaverse in UAE is taken into consideration | Only personal innovativeness and user satisfaction is taken into account. Perceived Ease of Use and Perceived Usefulness only were considered | Focus to be given on other medical aspects as well |
| [158] | Virtual Reality | Immersion, collaboration and interaction could be greatly improved with the intervention of the Metaverse in onlile pedagogy | Only small group size was considered for evaluation | To understand the detrimental effect of adoption of metaverse in medical education |
| [159] | Augmented Reality, Virtual Reality | To provide counselling services to post-operative patients | Set up needs to be changed for addressing a large group of patients | Can be adopted for ICU |
| [160] | Wearable devices, IoT | A technique for providing social skills training for children affected with Autism Spectrum Disorder is proposed | Obtaining consent from guardians | Children of all categories can be considered for the study |
| [161] | Artificial Intelligence, Virtual Reality, Robotics | Adoption of the Metaverse in spine care with respect to education, diagnosis, consultation, surgery and research | Affordable advanced care facilities | Other advanced technologies can be incorporated |

Apart from the aforementioned projects, key players such as Google LLC, CableLabs, Roblox, Epic Games and Meta Platforms are already into research and development of the Metaverse Healthcare solutions [174]. Meanwhile, countries like Dubai have introduced Virtual Assets Law and have formulated Virtual Assets Regulatory Authority (VARA) in early 2022. Dubai's key the Metaverse initiatives are in the cryptocurrency and blockchain sectors [175].

## VI. CHALLENGES & OPEN ISSUES

The present technology trends in the healthcare industry have adopted remote patient monitoring in an efficient and effective manner, ensuring the best service to the patients using modern telemedicine techniques. The Metaverse is a new buzzword, expecting disruptive transformations in several aspects of life. However, the adoption of the Metaverse in the healthcare domain will boost the present patient health monitoring service by changing the way of interacting with healthcare systems with additional interactive features in a virtual world using technologies like virtual reality for medical training, augmented reality in surgical procedures, etc. Although the Metaverse provides a promising solution for the healthcare domain, it still suffers from some challenges such as

### A. DATA PRIVACY CONCERNS

With the development of communications and virtual technologies, the Metaverse aims to provide an excellent user experience for the patient in a virtual world by interconnecting the physical world to the virtual world. The Metaverse will manage and monitor the patient's physiological responses and body movements. Moreover, the Metaverse has the ability to collect personal information such as brainwaves, bio-metric data, health information, and preferences of individual patients while interacting with various services. The collected sensitive information is not safe in the virtual world, malicious users can imitate other users to obtain the personal information of the individuals or it may be leaked to third parties [176]. Considering the risk of private data leakage of users, one of the possible solution can be creating "clone cloud" and "private copy" to ensure the user's location information, activities, and goals, are preserved by confusing the intruder [177]. Maintaining privacy at the sensory level, the privacy of communication and behavior is very much needed in the existing the Metaverse-based healthcare solutions [178].

**TABLE 3.** Objectives of different the Metaverse healthcare projects

| Project | Country | Focus Area | | | | |
|---|---|---|---|---|---|---|
| | | Mental Health | Physical Health | Surgeries | Medical Education/Training | Others |
| Healthify | India | ✓ | ✓ | | | ✓ |
| Bump Galaxy | Netherlands | ✓ | | | | |
| AccuVein | USA | | | ✓ | | |
| HintVR | USA | | ✓ | ✓ | ✓ | ✓ |
| HealthBlocks | Netherlands | ✓ | ✓ | | | ✓ |
| Mesh | USA | | | | ✓ | |
| HoloLens 2 | USA | | ✓ | | ✓ | |
| IRIS | USA | | | ✓ | ✓ | |
| GHA Project | India+USA | | | | ✓ | |

### B. INFORMATION SECURITY CONCERNS

The giant healthcare organizations will collect sensitive information, with or without the concerns of the patients in various ways, while providing health services. With the aid of distinct communication and virtual technologies, the Metaverse would be able to provide an excellent virtual experience for doctors to treat patients with various health disorders remotely with help of various electronic gadgets, generating an enormous amount of digital data. Since the Metaverse is a blend of modern technologies, during the process, the patients' current health conditions are transferred through the communication channel and the doctor's responses are also transferred through the same channel. However, electronically generated medical data consists of sensitive information related to patients, and securing these from external breaches or attackers at distinct levels of the process is a challenging task. The AR/VR devices are the entry point to malware invasions and data breaches [179].

### C. INTEROPERABILITY ISSUES

Healthcare interoperability depicts the capabilities of healthcare providers and other systems to electronically share patients' information without any hindrance i.e., the electronic health record system of one provider should be in a position to transfer the patient data to another provider's system with the help of modern technologies and electronic equipment ensuring consistency and availability of sensitive information related to various patients. As the number of health data and modernized devices in the healthcare domain increases, interoperability becomes a challenging task [180]. Adoption of the Metaverse in healthcare will open doors to new challenges by incorporating various hardware and software components with wearable equipment in a virtual environment. Every the Metaverse-based healthcare service should ensure the security and consistency of sensitive information by interacting through the various components, starting from data sensing to data processing. Proper communication standards and data adoption strategies need to be implemented to avoid overcoming the severe and unforeseen consequences while migrating the devices from traditional healthcare services to the virtual world [176].

### D. HIGH COST OF TECHNOLOGY

Due to new medical innovations, healthcare technology is constantly evolving leading to digital transformations by incorporating robotic surgeries and virtual reality in various medical services. The blend of augmented reality and virtual reality is mostly used in medical training and surgical procedures to perform complex surgeries with extreme precision. These software and hardware components will improve the performance of medical devices and equipment. The emergence of the Metaverse-focused companies has increased in development of advanced AR and VR-based solutions to improve the overall surgical environment in a global market. Transforming the present healthcare system with the Metaverse in an effective manner requires high-tech wearables like glasses, gloves, sensors, and other hardware components that can sense the state of the patients accurately [181]. However, the cost of wearables is very high, and to avail new capabilities, we will also need to buy new equipment in line with the development of technology. Since the Metaverse requires high end-to-end connectivity for an efficient operation, the cost of infrastructure will also be very high for the healthcare providers. At the outset, the adoption of the Metaverse in the healthcare domain, by transforming the traditional operations with modern wearable equipment, suitable software, and hardware infrastructure is very costly [182].

### E. THE PERSONAL TOUCH IS LOST

The modern digital transformation of healthcare systems with telemedicine service, and remote patient monitoring systems induces the gap between patient and doctor. Doctors will suggest the treatment remotely by hearing the details from patients or by looking at digital health records without actually looking into the patients. However, the the Metaverse will provide an immersive virtual world for patients and healthcare providers by providing effective, and efficient interactive options. Because of the modern digital mode of treatment, most of the patients feel lack the unique bonding of face-to-face rapport. The digital mode of treatment makes the patient feel lonely and also it affects the recovery rate [183].

## F. LEGAL AND REGULATORY CHALLENGES

Ever since the adoption of the Metaverse in the healthcare domain, it has transferred tremendously interns of providing service to the patients and educating the health professionals with modernized equipment and gadgets. The recent advancements in the market have transformed the Metaverse from technology into a new business model. As the practical applications of the Metaverse continue to grow with evolving technologies in the Healthcare domain, involving various entities such as Healthcare Providers, insurance, and pharmaceutical companies to provide efficient service to patients, may lead to legal and regulatory issues [184]. Since the Metaverse doesn't have proper standards and policies, its adoption may lead to confusion among various entities of the domain interns of Intellectual Property Rights, trust concerns,etc. However the proper policies need to be defined under the legal framework to take proper actions or to control the illegal operations in the virtual environment [88].

## VII. FUTURE DIRECTIONS

In this section we will discuss several future directions in the Metaverse for that are suitable for the healthcare domain.

### A. SECURITY EMPOWERED THE METAVERSE

The Metaverse is more prone to security threats because of the heterogeneous components involved in the patient healthcare system, providing security for user-sensitive information is the responsibility of the healthcare provider. The existing the Metaverse systems depend on frequent security patches upgrades to strengthen the deployed system. The security patches will enhance system security to some extent but the continuous cyber-physical attack on the Metaverse surface will lead to fragile patches. However, Endogeneous security such as quantum key distribution is a promising solution ensuring security by design mechanisms with self-protection, self-evolution, and autoimmunity capabilities considering security and privacy factors before the system design. Hence the future the Metaverse can resist the unknown and known security vulnerabilities and privacy threats [179].

### B. CLOUD-EDGE ORCHESTRATED THE METAVERSE

In the Metaverse, the traditional server-centric network architecture requires transmission of data between cloud and terminal devices, consuming huge bandwidth leading to high latency and data loss, which degrades the quality and the user experience. In the Metaverse, different users will have distinct QoS(Quality-of-Service) and QoE (Quality-of-Experience) requirements, based on the user requirements services need to the leveraged without any delay. However, the orchestrated edge-cloud the Metaverse architecture will ensure dynamically sharing computation, and communication among various entities by enhancing the QoE and QoS of various services. The self-balancing federated learning-based the Metaverse framework will resolve the imbalance in the distribution of computing power of edge nodes by enabling edge nodes to play the role of intermediaries which improves the privacy of the patients. However, future works need to be investigated including the design of edge–edge, edge-cloud, and edge-end collaboration mechanisms to strengthen the security and privacy aspects of a the Metaverse in the healthcare domain [179]. .

### C. ENERGY-EFFICIENT THE METAVERSE

the Metaverse is a blend of various technologies such as AR and VR which helps in creating a virtual world, the end-users are associated with various wearable devices to interact with each other. These wearables can be resource-constrained but their communication and computation capabilities with the heterogeneous components of the system make them more resource hungry and consume huge power leading to greenhouse gas emission, increasing environmental concerns. However, the development of a new the Metaverse architecture design with green edge-cloud computing, an energy-efficient consensus protocol to support green networking and computing in the Metaverse for healthcare [179].

### CONCLUSION

This paper presents an exhaustive survey of the applications of the Metaverse in healthcare. Firstly, the current digital healthcare scenario and the need for implementing the Metaverse for healthcare are briefed. Then, the existing state of the art technologies empowering digital and smart healthcare frameworks are discussed, followed by the discussion of the enabling technologies of the Metaverse. Later, the potential applications of the Metaverse for healthcare are discussed explicitly. To be more specific, the prospective implementation of the Metaverse in medical diagnosis, patient monitoring, healthcare training, surgeries, medical therapeutics and theranostics are emphasized. Additionally, the recent and upcoming projects relevant to the Metaverse for healthcare involving blockchain, digital twins and telemedicines are highlighted. The challenges and open issues for realizing the full potential of the Metaverse for healthcare are critically analysed and relevant future scope of directions are recommended that would enable confident implementation of the Metaverse in healthcare.

**TABLE 4.** Summary of Challenges and Future directions.

| Challenge | Description | Possible Solutions | Future Directions |
|---|---|---|---|
| Cybersecurity Risks | - The digital data collected from various gadgets are exposed to hackers or attackers in various stages of communication. | - The hardware and software components used in the system must be in a position to detect risk or attacks<br>- Defining the strict communication rules between hardware's in the Metaverse network.<br>- Adopting Zero Trust Policy | - Healthcare organizations can have cyber security protocols with strict measures such as network slicing with ring-fencing<br>- Adoption of Lightweight Blockchain technology with optimized cryptography techniques and consensus algorithms for securing the user data |
| Privacy Risks | - Safeguarding and preserving the virtual identities of the patients<br>- Incorporating modern technologies to develop Virtual environment in healthcare domain like XR, AI, IoT will expose user sensitive data to the outer world, hence privacy breaches will happen. | - To ensure the privacy of user sensitive data, throughout the data cycle in virtual environment can be achieved by implementing application-layer encryption.<br>- Adaptation of privacy frameworks by allowing users and developers to fine control the privacy of input data.<br>- Adopting privacy enhancing technologies (PETs) to blur sensitive data collected from the sensors before being shared with cloud services. | - Adopting Federated learning techniques for enhanced privacy of patient's sensitive data. |
| Delay and Communication Challenges | - Establishing proper communication among heterogeneous devices in the Metaverse environment ensuring there will be no delay, no data loss is a challenging task.<br>- The Metaverse requires high speed communication channel for data transfer | - Designing an Auction based Deep reinforcement learning to improve communication efficiency.<br>- Implementing Incentive based mechanism in Blockchain technology. | - Adopting 6G networks and service models for providing efficient service using the Metaverse in healthcare. |
| Computational and Predictability Challenges | - Even though the existing AI-enabled solutions for the Metaverse sharpen the QoS, but requires massive computation capabilities. | - Designing optimized, efficient AI Models<br>- Deploying AI at Edge | - Lightweight AI models for better prediction and reduces computational<br>- Adopting Explainable AI (XAI) in the Metaverse for improving prediction by understanding AI models efficiently. |
| High Cost | - The Metaverse uses combination of various technologies to enable the immersive 3D environment, the avatars need various gadgets for interacting with the virtual environment.<br>- The deployment, maintenance and cost of various interacting gadgets will be costly. Hence the overall infrastructure cost will be very high. | - Developing devices with cost efficient and more durable.<br>- Adopting the decentralized cloud storage system for storing the data. - Adopting off-chain blockchain Models for reducing the storage and computational cost. | - Developing the Metaverse by adopting Web3.0 infrastructure rather than Web2.0. |
| Standards and Ethical issues | - No proper Standards are defined for Healthcare the Metaverse. Hence, each and every healthcare provider has to create their own the Metaverse environment which is expensive and time-consuming. | - Adopting Contextual Approach i.e, an act of investigating context of technology and determining the constraints on the flow of information.<br>- Other than the existing govt standards a separate regulatory framework for balancing the rights of users and the growth of technology. | - Adopting Consumer-centric approach<br>- Implementing Walled gardens – no company or govt should be owing the Metaverse it should work like internet and provide gateways into walled gardens |

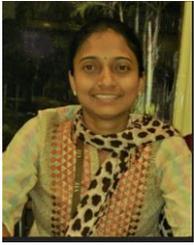
RAJESWARI CHENGODEN is currently working as an Associate Professor in the School of Information Technology and Engineering, Vellore Institute of Technology, Vellore, Tamil Nadu, India. She obtained her Bachelors in Computer Science and Engineering, Masters in Software Engineering and Ph.D. in Information and Communication Engineering, in the year 2006, 2008, and 2017 respectively from Anna University, Chennai, Tamil Nadu, India. She has more than 12 years of experience in teaching. Currently, her areas of research include Machine Learning, Internet of Things, Deep Neural Networks, Blockchain and Machine Fault Diagnosis.

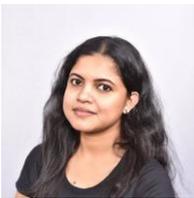
NANCY VICTOR is an Assistant Professor Senior in the School of Information Technology & Engineering at the Vellore Institute of Technology, Tamil Nadu, India, where she has been a faculty member since 2013. Prior to this, she has worked as an Assistant Professor at MCET, Kerala, India. Nancy has a decade of experience in the teaching field. She is a member of LITD'27, Bureau of Indian Standards. She is also a member of Indian Science Congress and life member of Indian Society for Technical Education. She received her PhD from Vellore Institute of Technology (2021), M.Tech degree in Computer and Information Technology (CIT) from Manonmaniam Sundaranar University (2012), and her B.Tech in Information Technology from Kerala University (2010). She has received several recognitions including the Gold Medal from the then Governor Dr. K. Rosaiah for securing University First Rank in M.Tech- CIT. She has also been selected for the "Advance Training for Professionals", fully funded by Atal Bihari Vajpayee Indian Institute of Information Technology and Management, Gwalior, to meet the immediate trained human resource requirements of IT and ITES industry. Nancy has published her research works in various International Conferences and journals. Her research spans the fields of Deep Learning, Natural Language Processing, Data Privacy and Big Data.

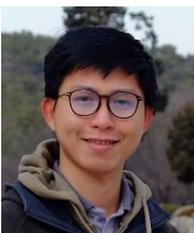
THIEN HUYNH-THE (Member, IEEE) received the B.S. degree in electronics and telecommunication engineering from Ho Chi Minh City University of Technology and Education, Vietnam, in 2011, and the Ph.D. degree in computer science and engineering from Kyung Hee University (KHU), South Korea, in 2018. From March to August 2018, he was a Postdoctoral Researcher with KHU. He is currently a Postdoctoral Research Fellow with ICT Convergence Research Center, Kumoh National Institute of Technology, South Korea. He is awarded with the Superior Thesis Prize by KHU in 2018 and the Golden Globe Award for Vietnamese young scientists by Ministry of Science and Technology of Vietnam in 2020. His current research interest includes radio signal processing, digital image processing, computer vision, wireless communications, machine learning, and deep learning.

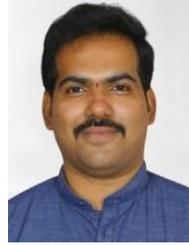
GOKUL YENDURI received his Master's degree (M.Tech, IT) from Vellore Institute of Technology in the year 2013 . Currently, he is a Research Scholar at Vellore Institute of Technology, Vellore, India. His areas of interest are in machine learning, and predictive analysis, software engineering, computer networks, network security. He has worked as an Assistant Professor in past. He attended several National, International Conferences, Workshops, Guest Lectures, and published papers in peer-reviewed international journals. He also carried out several academic responsibilities like a key member in several accreditation committees.

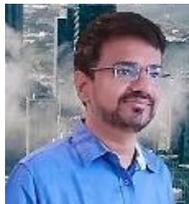
RUTVIJ H. JHAVERI (Senior Member, IEEE) is an experienced educator and researcher working in the Department of Computer Science & Engineering, Pandit Deendayal Energy University, Gandhinagar, India. He conducted his Postdoctoral Research at Delta-NTU Corporate Lab for Cyber-Physical Systems, Nanyang Technological University, Singapore. He completed his PhD in Computer Engineering in 2016. In 2017, he was awarded with prestigious Pedagogical Innovation Award by Gujarat Technological University. Currently, he is co-investigating a funded project from GUJCOST. He was ranked among top 2% scientists around the world in 2021. He has 2100+ Google Scholar citations with h-index 23. Apart from serving as an editor/ guest editor in various journals of repute, he also serves as a reviewer in several international journals and also as an advisory/TPC member in renowned international conferences. He authored 120+ articles including the IEEE/ACM Transactions and leading IEEE/ACM conferences. Moreover, he has several national and international patents and, copyrights to his name. He also possesses memberships of various technical bodies such as ACM, CSI, ISTE, IDES and others. He is a member of the Advisory Board in Symbiosis Institute of Digital and Telecom Management, Manav Rachna Group and Sandip University since 2022. He is an editorial board member in several Hindawi and Springer journals. He also served as a committee member in "Smart Village Project" - Government of Gujarat, at the district level during the year 2017. His research interests are SDN, network security/resilience, IoT systems and eHealth.

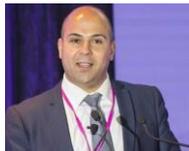
MAMOUN ALAZAB (Senior Member, IEEE) received the Ph.D. degree in computer science from the School of Science, Information Technology and Engineering, Federation University of Australia. He is currently an Associate Professor with the College of Engineering, IT, and Environment, Charles Darwin University, Australia. He is also a Cyber Security Researcher and a Practitioner with industry and academic experience. His research is multidisciplinary that focuses on cyber security and digital forensics of computer systems with a focus on cybercrime detection and prevention. He has more than 150 research articles in many international journals and conferences, he delivered many invited and keynote speeches, 24 events in 2019 alone. He convened and chaired more than 50 conferences and workshops. He also works closely with government and industry on many projects, including Northern Territory (NT) Department of Information and Corporate Services, IBM, Trend Micro, the Australian Federal Police (AFP), the Australian Communications and Media Authority (ACMA), Westpac, and United Nations Office on Drugs and Crime (UNODC). He is also the Founding Chair of the IEEE NT Subsection.

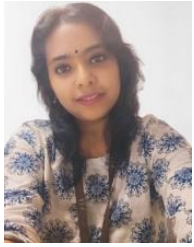

SWETA BHATTACHARYA is currently associated with Vellore Institute of Technology (University), as an Assistant Professor in the School of Information Technology & Engineering. She has received her Ph.D. degree from Vellore Institute of Technology and Master's degree in Industrial and Systems Engineering from State University of New York, Binghamton, USA. She has guided various UG and PG projects and published peer reviewed research articles. She is also a member of the Computer Society of India and Indian Science Congress. Her research experience includes working on Pill Dispensing Robotic Projects, as a fully funded Watson Research Scholar at Innovation Associates, Binghamton at SUNY. She has completed six sigma green belt certification from Dartmouth College, Hanover. Her research interests include applications of machine learning algorithm, data mining, simulation and modeling, applied statistics, quality assurance and project management.

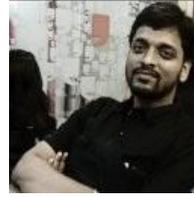

THIPPA REDDY GADEKALLU is currently working as an Associate Professor in the School of Information Technology and Engineering, Vellore Institute of Technology, Vellore, Tamil Nadu, India. He obtained his Bachelors in Computer Science and Engineering from Nagarjuna University, India, in the year 2003, Masters in Computer Science and Engineering from Anna University, Chennai, Tamil Nadu, India in the year 2011 and his Ph.D in Vellore Institute of Technology, Vellore, Tamil Nadu, India in the year 2017. He has more than 14 years of experience in teaching. He has more than 150 international/national publications in reputed journals and conferences. Currently, his areas of research include Machine Learning, Internet of Things, Deep Neural Networks, Blockchain, Computer Vision. He is an editor in several publishers like Springer, Hindawi, Plosone, Scientific Reports (Nature), Wiley. He also acted as a guest editor in several reputed publishers like IEEE, Elsevier, Springer, Hindawi, MDPI. He is recently recognized as one among the top 2% scientists in the world as per the survey conducted by Elsevier in the year 2021.

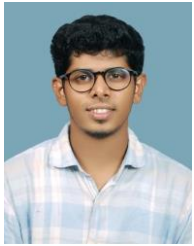

PAWAN HEGDE received his Bachelor of Engineering(B.E, CS) from VTU Belgaum in 2010 and Master's degree (M.Tech, CS) from Jain University Bangalore in the year 2012 . Currently, he is a Research Scholar at Vellore Institute of Technology, Vellore, India. His areas of interest are in Blockchain Technology, IoT security, machine learning and network security. He has worked as an Assistant Professor at NMAMIT Nitte. He attended several National, International Conferences, Workshops,and published papers in peer-reviewed international journal

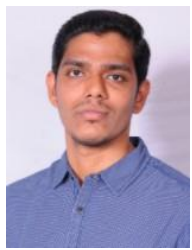

PRAVEEN KUMAR REDDY MADDIKUNTA is currently working as Assistant Professor Senior in the School of Information Technology and Engineering, VIT, Vellore, Tamil Nadu, India. He obtained his B.Tech. in CSE from Jawaharlal Nehru Technological University, India, M.Tech. in CSE from VIT University, Vellore, Tamil Nadu, India, and completed his Ph.D. in VIT, Vellore, Tamil Nadu, India. He was a Visiting Professor with the Guangdong University of Technology, China, in 2019. He worked with IBM, Alcatel-Lucent as a software developer in 2011 and 2013. He produced more than 100 international/national publications.